\pdfoutput=1

\documentclass[11pt]{article}

\usepackage[final]{acl}

\usepackage{times}
\usepackage{latexsym}

\usepackage[T1]{fontenc}

\usepackage[utf8]{inputenc}

\usepackage{microtype}

\usepackage{inconsolata}

\usepackage{graphicx}
\usepackage{amssymb}
\usepackage{amsmath}

\usepackage{amsmath,amsfonts,bm}

\def\eqref#1{equation~\ref{#1}}

\def\1{\bm{1}}

\DeclareMathAlphabet{\mathsfit}{\encodingdefault}{\sfdefault}{m}{sl}
\SetMathAlphabet{\mathsfit}{bold}{\encodingdefault}{\sfdefault}{bx}{n}

\usepackage{booktabs}
\usepackage{multirow}
\usepackage{enumitem}
\usepackage{float}

\usepackage{subcaption}
\usepackage{cleveref}
\usepackage{tabularx}

\usepackage[table]{xcolor}
\definecolor{darkblue}{rgb}{0.0,0.0,0.65}
\definecolor{darkred}{rgb}{0.65,0.0,0.0}
\definecolor{darkgreen}{rgb}{0.0,0.5,0.0}
\definecolor{tab:blue}{RGB}{31,119,180}  
\definecolor{tab:red}{RGB}{214,39,40}  
\definecolor{tab:green}{RGB}{44,160,44}  
\definecolor{tab:orange}{RGB}{255,127,14}  

\hypersetup{
	colorlinks = true,
	citecolor  = darkblue,
	linkcolor  = darkred,
	filecolor  = darkblue,
	urlcolor   = darkgreen,
}
\title{Quantifying Aleatoric Uncertainty of In-Context Learning\\for Robust Measure of LLM Prediction Confidence}

\author{
  Jinseok Chung \quad
  Minkyoung Song\textsuperscript{*} \quad
  Hyunji Jung\textsuperscript{*} \quad
  Namhoon Lee \\
  POSTECH \\
  \texttt{\{jinseok.chung, minkyoung.song, hyunji.jung, namhoon.lee\}@postech.ac.kr}
}

\begin{document}

\maketitle
\begingroup
\renewcommand\thefootnote{\fnsymbol{footnote}}
\footnotetext[1]{Equal contribution.}
\endgroup
\begin{abstract}
In-Context Learning (ICL) allows LLMs to adapt to new tasks from a few demonstrations, but its reliability remains a concern: predictions are highly sensitive to both prompt design and the model's ability to understand the context, obscuring whether failures arise from data properties or model limitations.
Uncertainty decomposition—separating aleatoric from epistemic sources—is particularly crucial in this setting, yet existing methods, designed for standard generation tasks, fail to capture the unique dynamics of ICL.
To address this, we introduce a concept of \textit{self-function vectors}, built upon Bayesian views and the mechanistic interpretability of ICL.
These vectors leverage internal model representations to model the latent concept learned during in-context prompting, thereby enabling a direct estimation of aleatoric uncertainty within a Bayesian framework and circumventing the reliance on brittle input or decoding manipulations.
Given the lack of established benchmarks and suitable evaluation protocols, we also propose the first and rigorous evaluation protocol, in which data is manipulated in controlled ways so as to quantify aleatoric uncertainty precisely and separately from epistemic uncertainty.
With this new evaluation framework, initially grounded in synthetic tasks for conceptual development and subsequently extended to real-world datasets, we show that our proposed methodology can measure uncertainty of LLM predictions made under ICL more reliably than existing alternative methods.
Moreover, we show it can be used as a practical tool for trustworthy-related applications, such as hallucination detection.
Our findings pave a new direction for connecting the quantitative view of uncertainty with the mechanistic understanding of model behavior.
\end{abstract}
\section{Introduction}

\begin{figure*}[!t]
    \centering
    \includegraphics[width=0.85\linewidth]{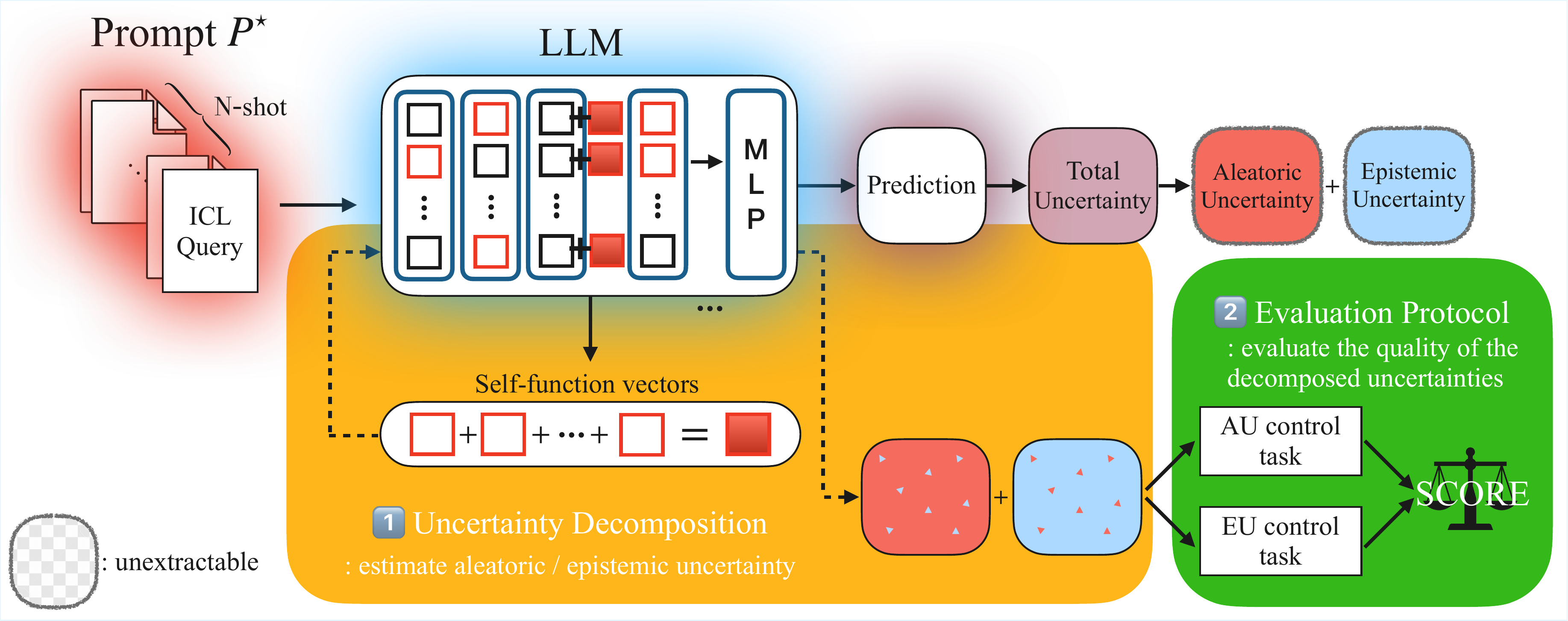}
    \caption{
    This work comprises two main pillars. (1) Uncertainty Decomposition: We probe the LLM's internal representations to construct self-function vectors from the activations of salient attention heads. These vectors serve as a proxy for the latent concept learned from the prompt, enabling a principled estimation of AU and 
    EU. (2) Evaluation Protocol: We then assess the quality of the decomposition using a novel framework with tasks specifically designed to independently perturb AU and EU, thereby providing a robust score for decomposition fidelity.}
    \label{fig:overview}
\end{figure*}

In-Context Learning (ICL) enables Large Language Models (LLMs) to adapt to new tasks without parameter updates by conditioning on a handful of input–output demonstrations provided directly in the prompt \citep{Brown_fewshot}. 
Despite its versatility, ICL is inherently brittle: predictions can fluctuate with small changes to prompt format, the ordering of demonstrations, or data properties \citep{min2022rethinkingroledemonstrationsmakes, NEURIPS2022_77c6ccac, chan2025toward}. 
On top of that, ICL predictions depend highly on the model’s context learnability \citep{wang2024twostage}, making it difficult to discern whether failures stem from data properties or from model limitations.
This highlights the importance of not only measuring the overall level of uncertainty—known as Uncertainty Quantification \citep{kuhn2023semantic, shorinwa2024surveyuncertaintyquantificationlarge}—but also disentangling its underlying sources, a process referred to as Uncertainty Decomposition \citep{Kendall2017, abbasi-yadkori2024to, wang2025on, jayasekera2025variationaluncertaintydecompositionincontext}.

Uncertainty decomposition aims to disentangle uncertainty into its principal components: aleatoric uncertainty (AU), which stems from inherent data ambiguity, and epistemic uncertainty (EU), which reflects limitations in the model’s knowledge or representations.
This decomposition offers sharper insight into failure modes: AU reflects irreducible noise, while EU highlights uncertainty that could, in principle, be reduced with better modeling or more evidence.
This distinction is especially important in ICL, where predictions hinge on both the given context and the model’s learnability.

Yet, current research is primarily focused on standard settings such as Question and Answer, and not much has been done for ICL \citep{shorinwa2024surveyuncertaintyquantificationlarge}.
Prior approaches typically involve manipulating inputs \citep{abbasi-yadkori2024to, ling-etal-2024-uncertainty, wang2025on} or altering decoding strategies \citep{kuhn2023semantic, ling-etal-2024-uncertainty}, but such methods do not readily transfer to ICL, as seemingly small variations in inputs or decoding may lead to substantial and unpredictable fluctuations in performance \citep{min2022rethinkingroledemonstrationsmakes}.

Encouragingly, two recent lines of research suggest a path forward.
First, interpreting ICL as implicit Bayesian inference provides a natural theoretical foundation for principled decomposition \citep{xie2022an, wang2023large, jiang2024a}.
Second, advances in the mechanistic interpretability—examining model internals such as representation dynamics and head activations—offer tools to observe ICL behavior without requiring explicit manipulations of inputs or decoding \citep{hendel2023incontext, todd2024function, heo2025do}.
Taken together, these perspectives provide a practical tool for principled and ICL-specific uncertainty decomposition.

At the same time, evaluating decomposition in ICL remains an open challenge: existing studies typically rely on downstream tasks from uncertainty quantification—such as hallucination detection or out-of-distribution identification —since no benchmark directly assesses decomposition quality, leaving it unclear whether AU and EU are genuinely separated or whether apparent gains merely reflect proxy-task performance.

Our contributions are summarized as follows:  
\begin{enumerate}  
    \item We introduce an aleatoric uncertainty quantification methodology based on \textit{self-function vectors}—a novel approach that leverages internal model representations for principled uncertainty decomposition in ICL, building upon Bayesian perspectives and mechanistic interpretability.
    \item We develop an evaluation protocol tailored to uncertainty decomposition in the ICL setting, grounded in synthetic tasks that can reliably capture ICL dynamics and enable faithful assessment of decomposition quality.
    \item We further establish the applicability of our method by showing competitive performance on hallucination detection. 
    This opens a new potential for bridging uncertainty estimation with mechanistic interpretability.
\end{enumerate}

\section{Background}
LLMs have demonstrated the ability to ``learn'' new tasks without updating their weights by conditioning on a small number of labeled examples provided on the prompt. 
This behavior, known as In-Context Learning(ICL), allows the model to condition its predictions on a few labeled examples provided at inference time—without any model updates.
Formally, let $\mathcal{D}_{\text{ex}} = \{(x_i, y_i)\}_{i=1}^{k}\subseteq \mathcal{D}_T$ denote the example set sampled from the dataset $\mathcal{D}_T$ of task $T$, and let $x^\star$ be the query.
Conditioned on $\mathcal{D}_{\text{ex}}$, a model defines the predictive distribution:
\begin{align}
    p\bigl(y \mid x^\star, \mathcal{D}_{\text{ex}}\bigr).
\end{align}
All task adaptation occurs within the context window, with no updates to model weights. 
This surprising behavior has inspired research to understand its underlying principles. 
The most prominent perspective is Bayesian inference, which views ICL as posterior updating on the observed data.

\paragraph{Bayesian View of ICL} 
A widely held view is that ICL corresponds to \emph{implicit Bayesian inference} \citep{xie2022an,wang2023large, jiang2024a}. In this perspective, the model infers a latent concept $\phi$ from the prompt and conditions its predictions on that inferred belief. 
The resulting predictive distribution approximates
\begin{align}
 p\bigl(y &\mid x^\star, \mathcal{D}_{\text{ex}}\bigr)\nonumber\\
& \approx \int p(y \mid x^\star, \phi)\;\underbrace{p(\phi \mid \mathcal{D}_{\text{ex}})}_{\text{implicit posterior}}\, d\phi.
\end{align}
Accordingly, the total uncertainty can be decomposed information-theoretically
\begin{align}
    \mathcal{H}[p(y \mid x^\star, \mathcal{D}_{\text{ex}})&]= \underbrace{\mathcal{I}(y; \phi \mid x^\star, \mathcal D_{\text{ex}})}_{\text{Epistemic}}\\ & + \underbrace{\mathbb{E}_{p(\phi \mid \mathcal D_{\text{ex}})}[\mathcal{H}(p(y \mid x^\star, \phi))]}_{\text{Aleatoric}},\nonumber
\end{align}
where $\mathcal{H}$ and $ \mathcal{I}$ denote the Shannon entropy and mutual information, respectively. \citep{pmlr-v216-wimmer23a, falck2024is, jayasekera2025variationaluncertaintydecompositionincontext}.

\paragraph{Mechanistic View of ICL} 
Mechanistic interpretability is a line of research that seeks to understand how models operate by identifying the specific neurons or attention heads that contribute to their behavior and how they work together to produce the model’s output.
Applied to ICL, this perspective aims to uncover how and which internal representations of the model process understanding context.
Early studies highlight \emph{induction heads}—attention heads that copy earlier tokens to later positions—as central to pattern completion \citep{elhage2021mathematical, olsson2022incontextlearninginductionheads}. 
More recent work identifies a complementary mechanism: \emph{function vectors}, compact task representations encoded in the activations of specific attention heads during ICL \citep{todd2024function}. 
An function vector $v_T$ is constructed by summing the prompt-wise averaged activations across all attention heads,
\begin{align}\label{eqn:fv}
    &\bar{h}^{T}_{(\ell,k)} = \frac{1}{n} \sum_{i = 1}^{n} h^{P_i}_{(\ell,k)},  \\
    &v_T = 
    \sum_{(\ell,k) \in \mathcal{S}_T} \bar{h}^{T}_{(\ell,k)},
\end{align}
where $\mathcal{S}_T$ denotes the set of causal heads (i.e., the attention heads most critical for task $T$), while $\bar{h}^{T}_{(\ell,k)}$ is defined as the average activation of the $k$-th head in layer $\ell$ computed over the input prompts $P_i$.

A function vector $v_T$ can be utilized by injecting it into the model’s hidden states—that is, by adding the vector to the hidden representations during inference.
Injecting $v_T$ shifts the context of the model's predictions—even in zero-shot settings—indicating that function vectors encode task identity and serve as internal task representations.
Building on this view, \citet{jiang2025unlocking} show that function vectors serve as latent task embeddings that parameterize task-specific functions, i.e., conditioning on $x^\star$ and $v_T$ induces behavior consistent with $p(y \mid x^\star, \mathcal{D}_{\text{ex}})$. 
Complementary ablations by \citet{yin2025attentionheadsmatterincontext} further reveal that function vector heads, rather than induction heads, are the dominant contributors to ICL performance in larger models.
Collectively, these findings establish function vectors as the primary carriers of task structure.

\paragraph{Uncertainty Decomposition before ICL}
The decomposition of predictive uncertainty into epistemic and aleatoric components has been explored widely before ICL. 
In computer vision, uncertainty decomposition underpins calibration, segmentation, and out-of-distribution detection, with EU flagging model limitations \citep{Kendall2017}. 
In NLP, it informs machine translation and text generation, separating hallucinations (epistemic) from linguistic ambiguity (aleatoric) \citep{kuhn2023semantic}. 
Active learning also exploits this distinction: EU drives sample acquisition, while AU marks uninformative or noisy data \citep{houlsby2011bayesianactivelearningclassification, Gal2017}.
Despite these advances, evaluating decomposition remains challenging: naive metrics often conflate the two sources, motivating principled benchmarks and task-specific frameworks \citep{mucsanyi2024benchmarking, smith2025rethinking}.
This evaluation gap is even more pronounced in ICL, where adaptation occurs entirely within the context window, entangling data- and model-driven variability in ways absent from conventional training regimes.
\section{Method}
We introduce a method for quantifying AU in ICL by leveraging the activations of task-specific attention heads.

Central to this approach is the novel concept of \textit{self-function vectors}, a variant of function vectors, designed to serve as proxies for latent concepts.
The method proceeds in four stages:
\begin{enumerate}[leftmargin=*, label=\textbf{S\arabic*.}]
    \item \textbf{Causal head selection:} 
    Identify salient attention heads $\mathcal{S}_T$ that cause the contextualization of $\mathcal{D}_T$ via causal indirect effect analysis (\Cref{method:identify_heads}).
    \item \textbf{Self-function vector construction:} 
    For a given prompt $P^\star=[\mathcal{D}_{\text{ex}}, x^\star]$, extract final-token activations $a_{(\ell,k)}$ from each $(\ell,k) \in \mathcal{S}_T$ and construct a self-function vector $\hat{v}_T$ (\Cref{method:extract_activations}).
    \item \textbf{Self-function vector intervention:} 
    Inject $\hat{v}_T$ into the hidden state at the final token during inference, yielding a latent-conditioned prediction $p(y \mid x^\star, \mathcal{D}_{\text{ex}}, \hat{v}_T)$. (\Cref{method:intervention}).
    \item \textbf{Uncertainty decomposition:} 
    Compute decomposed uncertainties using latent-conditioned predictions (\Cref{method:quantification}). 
\end{enumerate}
Unlike conventional approaches that infer uncertainty only from output variability (e.g., decoding or prompting strategies), our method probes the model’s internal representations.
This yields a structure-aware and interpretable foundation for uncertainty estimation in ICL \citep{shorinwa2024surveyuncertaintyquantificationlarge}.
Our code is available at \url{https://github.com/LOG-postech/self-fv-icl}.

\subsection{Identifying Salient Attention Heads}\label{method:identify_heads}
To identify task-relevant attention heads, we begin with a given task dataset $\mathcal{D}_T$ and estimate the causal contribution of each attention head to the model’s predictions. 
Following \citet{todd2024function}, we evaluate this by introducing counterfactual in-context inputs using label-shuffled examples $\tilde{\mathcal{D}}_{\text{ex}}$, which typically degrade performance. 
Given a corrupted prompt $\tilde{P}=[\tilde{\mathcal{D}}_{\text{ex}}, x_{val}]$, we assess each head $(l, k)$, by replacing its final-token activation during inference with the mean activation vector $\bar{h}_{(l,k)}^T$ computed over correctly labeled inputs. 
The causal effect (CE) of a head is then defined as the change in prediction probability induced by this replacement:
\begin{align}
\mathrm{CE}(P)=&p_{{h_{(\ell,k)}^{\tilde{P}} \rightarrow \bar{h}_{(\ell,k)}^T}}(y\!\mid[\tilde{\mathcal{D}}_{\text{ex}}, x_{val}])\nonumber\\&-p(y \mid\![\mathcal{D}_{\text{ex}}, x_{val}]).
\end{align}
The notation $\tilde{h}_{(\ell,k)}^T \rightarrow \bar{h}_{(\ell,k)}^T$ denotes the activation from a label-shuffled input is substituted with $\bar{h}_{(\ell,k)}^T$. 
We compute the causal effect multiple times using various prompts, and select the heads with the highest average CE to form the salient set $\mathcal{S}_T$, as they consistently encode task-relevant information \citep{jiang2025unlocking}.

\subsection{Self-Function Vectors as Latent Task Representations}\label{method:extract_activations}
Given a test prompt $P^\star$, we extract the final-token activations $h^{P_{\star}}_{(\ell,k)}$ from each salient head $(\ell,k) \in \mathcal{S}_T$. 
For each ensemble iteration $i$, we construct a function vector $\hat{v}_T^{(i)}$ by randomly sampling from $\{h_{(\ell,k)}^{P^\star}\mid (\ell,k) \in \mathcal{S}_T\}$, 
\begin{align}\label{eqn:self_fv}
    \hat{v}_T^{(i)} = \sum_{(\ell,k) \subseteq \mathcal{S}_T} h^{P^\star}_{(\ell,k)} .
\end{align}
We term these \emph{self-function vectors}, as they encapsulate the prompt’s internal representation through task-relevant attention heads.
This allows capturing prompt-specific uncertainty, rather than the averaged representation of the function vector.
This claim is supported by the experiments in \Cref{sec:experiments}.

\subsection{Self-Function Vectors Intervention}\label{method:intervention}
To obtain latent-conditioned predictions, we inject a self-function vector $\hat{v}_T^{(i)}$ into the hidden state at a designated target layer $\ell_t$ during inference, following \citet{todd2024function}. 
Let $h_{\ell_t}$ denote the original activation at layer $\ell_t$; the intervention modifies it as
\begin{align}
    h_{\ell_t}^{\prime} = h_{\ell_t} + \hat{v}_T^{(i)} .
\end{align}
The modified activation $h_{\ell_t}^{\prime}$ is then passed through the language model head to produce the predictive distribution
\begin{align}
    p(y \mid x^\star, \mathcal D_{\text{ex}}, \hat{v}_T^{(i)}),
\end{align}
as disccused in \citet{jiang2025unlocking}. 
This predictive distribution can be interpreted as a realization of $p(y \mid x^\star, \phi)$ with $\phi \sim p(\phi \mid \mathcal D_{\text{ex}})$. 
This interpretation naturally leads to the Bayesian decomposition of predictive uncertainty.

\subsection{Uncertainty Decomposition}\label{method:quantification}
We decompose predictive uncertainty using the standard Bayesian decomposition. 
Specifically, the total uncertainty is given by the entropy of the predictive distribution:
\begin{align}\label{eqn:total_entropy}
    \mathcal{H}&[p(y\mid x^\star, \mathcal D_{\text{ex}})]  \\ &= - \sum_{y \in \mathcal{Y}} p(y \mid x^\star, \mathcal D_{\text{ex}}) \log p(y \mid x^\star, \mathcal D_{\text{ex}})\nonumber.
\end{align}
AU is approximated by the average entropy of predictions under self-function vector interventions:
\begin{align}\label{eqn:intervene_entropy}
    \mathbb{E}&_{p(\phi \mid \mathcal D_{\text{ex}})}[\mathcal{H}(p(y \mid x^\star, \phi))]\nonumber\\
    &\approx\frac{1}{N} \sum_{i=1}^N \mathcal{H}(p(y \mid x^\star, \mathcal D_{\text{ex}}, \hat{v}_T^{(i)})).    
\end{align}
In practice, we instantiate \Cref{eqn:intervene_entropy} with $N=1$, performing a single intervention using a self-function vector $\hat{v}_T$ aggregated from the top causal-effect heads ($top\mbox{-}k\,\mathcal{S}_T$). Under this setup, AU is a deterministic quantity given a model and an input. We empirically compare this single top-$k$ intervention against ensemble variants that aggregate multiple interventions in \Cref{subsec:ensemble}, and find that the top-$k$ single intervention already captures the dominant effect, with ensembling yielding only marginal differences.
Finally, EU is obtained as the difference between the total and aleatoric terms.
\section{Evaluation Protocol}
Assessing the quality of uncertainty decomposition is critical for ensuring the reliability and interpretability of model predictions in ICL.
However, to the best of our knowledge, no evaluation protocol exists for this purpose.
To address this gap, we begin by analyzing how controlled perturbations to examples and queries affect uncertainty in a binary classification.
These simplified, well-controlled environments enable the precise attribution of changes in predictive uncertainty to either data- or model-induced factors.
Building on this analysis, we propose the first evaluation protocol specifically designed to assess the fidelity of uncertainty decomposition methods on language tasks.

\subsection{Toy Experiments}\label{subsec:synthetic_data}
A principled evaluation of uncertainty decomposition requires task settings in which the underlying sources of uncertainty can be manipulated independently.
Specifically, it should contain scenarios where AU increases while EU remains constant, and vice versa.

To identify perturbations that induce these distinct uncertainty dynamics, we design synthetic experiments that allow direct control over the data-generating process. This simplified setting offers two primary advantages: it mitigates uncontrolled sources of variation that could affect the uncertainties, and it enables the application of established decomposition techniques \cite{jayasekera2025variationaluncertaintydecompositionincontext}, which are theoretically well-grounded in such simplified settings.
The data distributions used in these experiments are visualized in \Cref{fig:synthetic_data_distr}.

\begin{figure}[!t]
    \centering
    \includegraphics[width=1.0\linewidth]{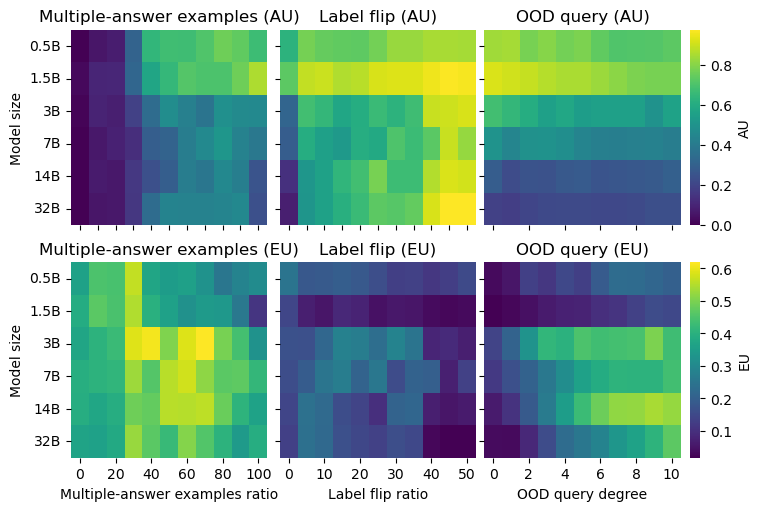}
    \caption{These results show how AU and EU change under various perturbation types and model sizes in synthetic tasks. 
    Each column corresponds to a specific perturbation setting.
    Within each subplot, the x-axis indicates the degree of perturbation, and the y-axis indicates model size. Color represents the magnitude of each uncertainty. 
    As model size increases, the Multiple-Answer Examples and Label Flip tasks control AU, while the OOD Query task controls EU.}
    \label{fig:summary_synthetic_exp}
\end{figure}

\Cref{fig:summary_synthetic_exp} shows the results of toy experiments, focusing on the tasks in which perturbations produced a clear separation between AU and EU dynamics. 

For large models, AU tends to increase as with more perturbation, whereas EU remains relatively unchanged for the Label Flip and Multi-answered Examples tasks as the perturbation degree increases.
In contrast, for the OOD Query task, EU increases with perturbation, whereas AU remains mostly fixed. 
This pattern aligns with prior results in computer vision, where AU is evaluated through human label disagreement and EU through capturing distributional shift \citep{mucsanyi2024benchmarking}.

Interestingly, smaller models exhibit changes in both AU and EU across all tasks. 
As model size increases, however, one source of uncertainty tends to stabilize while the other shows a more consistent increase—indicating improved separation of uncertainty sources in larger models.  

These results demonstrate that the designed perturbations can isolate distinct sources of uncertainty.
Further implementation details and results, including task variants where AU and EU cannot be clearly separated, are provided in \Cref{sec:toy_experiments}.

\subsection{WordNetMCQ-dataset Construction}
From the toy experiments, we identify three tasks controlling AU—Multiple-Answer Examples and Label Flip—and one task controlling EU—OOD Query. 
We then generalize these synthetic insights into tasks instantiated with WordNet \citep{miller-1994-wordnet}, thereby providing a more natural and language-grounded evaluation. 

\begin{figure}[!t]
    \centering
    \includegraphics[width=1.0\linewidth]{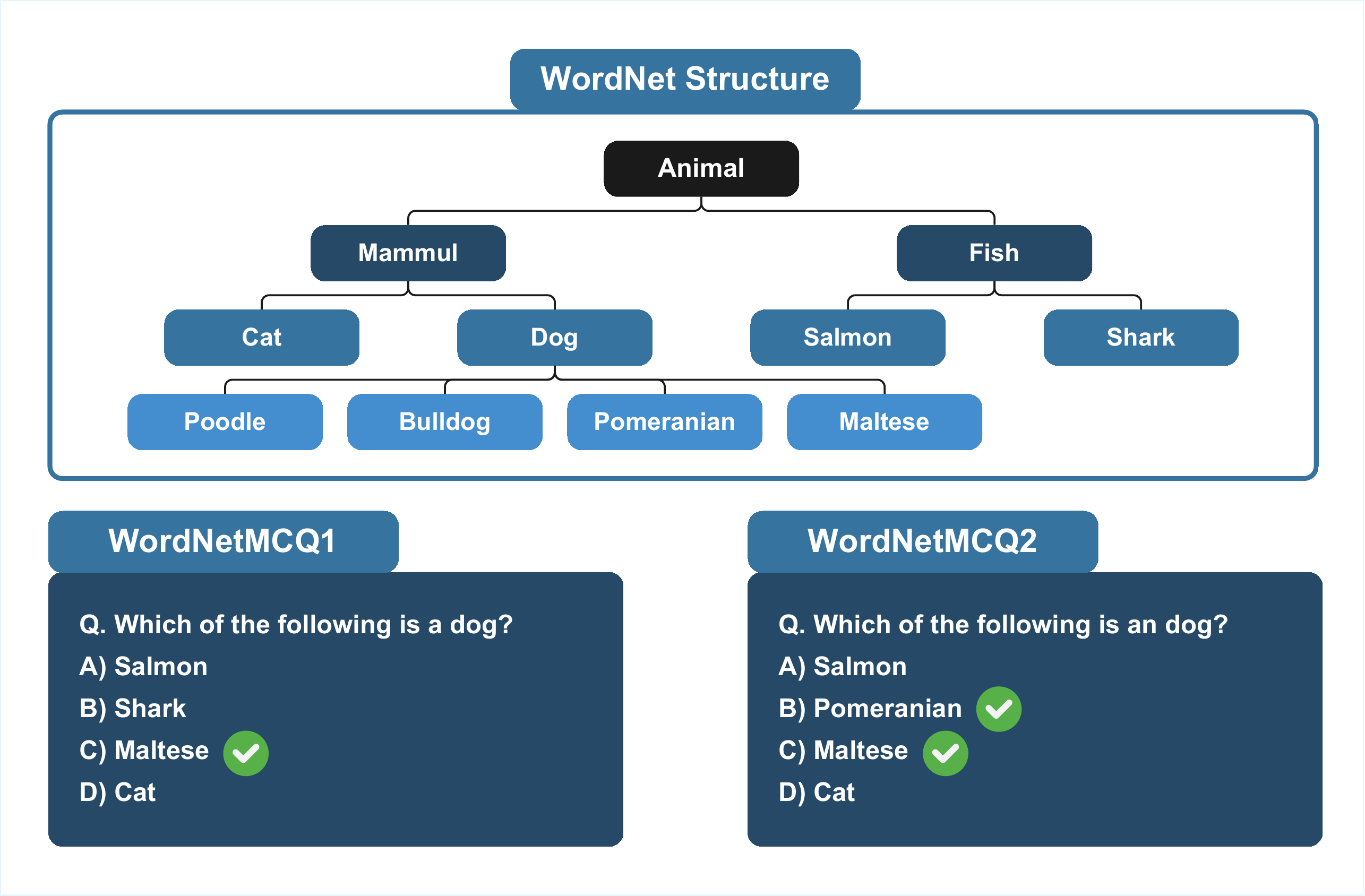}
    \caption{Construction of \textbf{WordNetMCQ1} (single-answer) and \textbf{WordNetMCQ2} (multiple-answer) question types from WordNet.
    This design provides the building blocks to carefully control task situations within ICL by varying the proportion of each question type in the demonstration prompts.}
    \label{fig:wordnet_structure}
\end{figure}

WordNet provides an ideal foundation, as it encodes fine-grained lexical semantics through synsets, hierarchical relations (e.g., hypernymy), and sense-level distinctions. 
Building on this, we construct multiple-choice questions, which we refer to as WordNetMCQ, following the subjective question setting \citep{abbasi-yadkori2024to}. Specifically, we distinguish between:
\begin{itemize}
    \item WordNetMCQ1: single-answer multiple-choice questions, where only one option is correct among four.
    \item WordNetMCQ2: multiple-answer multiple-choice question dataset in which each question has two correct answers.
\end{itemize}
This design allows us to carefully control situations for each task in ICL, as illustrated in \Cref{fig:wordnet_structure}.  

\paragraph{Multiple-Answer Examples} 
This task uses WordNetMCQ2 queries, while the examples are intentionally mixed between WordNetMCQ1 and WordNetMCQ2 instances. 
Consequently, if the examples contain mostly WordNetMCQ1 examples, the model tends to choose only one answer for a WordNetMCQ2 query. 
This shows how the ICL setup naturally varies the degree of ambiguity.

\paragraph{Label Flip} 
Here, we inject controlled label noise into the examples by flipping their labels according to a predefined ratio. 
As the flip ratio increases, the supervision becomes more inconsistent, forcing the model to learn from conflicting signals. 
This systematic perturbation directly raises AU, since the data itself becomes unreliable.  

\paragraph{OOD Query} 
To induce EU, we perturb the queries while preserving their original meaning. 
Specifically, we rephrase queries and insert special characters. 
By gradually increasing the intensity of these modifications, we adjust the degree of distributional shift for out-of-distribution inputs.

\subsection{Evaluation Metric}\label{subsec:evaluation_metric}
The degree of perturbation applied to each task is represented as an ordered categorical variable, whereas the resulting uncertainty is measured on a continuous scale.
Assessing the relationship between these two variables requires a correlation measure capable of capturing monotonic, and potentially non-linear, dependencies.

The statistical literature offers several such measures, including Pearson’s correlation coefficient, Kendall’s tau, and Somers’~$d$, each exhibiting different sensitivities to distributional characteristics.
Following the comparative framework proposed by \citet{GöktaşAtilaIsci}, we empirically evaluated these candidate measures.
Among them, Spearman’s rank correlation consistently demonstrated the most robust and reliable performance across our experimental conditions.

Based on these findings, we adopt Spearman’s rank correlation as our primary evaluation metric.
Details on the candidate measures, experimental design, and results are provided in \Cref{subsec:metric_details}.
\section{Experiments}\label{sec:experiments}
We evaluate our method through a series of uncertainty-controlled experiments and downstream benchmarks. 
The goals of our experiments are threefold: (i) to verify that our evaluation protocol can faithfully separate AU and EU in ICL, (ii) to demonstrate that the decomposed signals provide practical benefits in tasks such as hallucination detection, and (iii) to assess the specific contribution of self-function vectors through ablation analyses.

\paragraph{Experimental Setup}
We use LLaMA2-7B, LLaMA2-13B, and LLaMA2-70B \citep{touvron2023llama2openfoundation}, along with Qwen2.5-7B \citep{qwen2025qwen25technicalreport} and Mistral-7B \citep{jiang2023mistral7b}, as base models to assess robustness across both scale and architecture.
We restrict our method to the top-20 causal heads and intervene at one-third of the transformer's depth (e.g., the 10th layer for LLaMA2-7B, the 13th for LLaMA2-13B), following the layer-subsampling strategy \citep{todd2024function,liu2025IterativeVectors}.
This design avoids unrealistic full-layer searches. 
Evaluation primarily relies on WordNetMCQ, which enables controlled manipulation of uncertainty, with additional experiments on AG~News \citep{ag_news_citation}, Emotion \citep{emotion_citation}, HellaSwag \citep{hellaswag_citation}, and GSM8K \citep{gsm8k_citation} to assess applicability across classification, reasoning, and math domains (see \Cref{app:experiment_details} for further details).

\paragraph{Baseline} We consider Total entropy, Semantic Entropy \citep{kuhn2023semantic}, and UQ\_ICL \citep{ling-etal-2024-uncertainty} as representative baselines for uncertainty quantification and decomposition. 
Semantic Entropy measures uncertainty based on the entropy of semantically clustered model outputs, while UQ\_ICL decomposes total uncertainty in in-context learning into AU and EU.
We exclude methods that manipulate queries or vary the number of examples for estimation, as it is not straightforward to make direct comparisons with them. Extended baseline comparisons including MaxProb \citep{hendrycks2017baseline} and Lookback Lens \citep{chuang-etal-2024-lookback}, as well as a discussion of related work \citep{vazhentsev2025uncertaintyaware}, are provided in \Cref{app:additional_baselines}.

\subsection{Aleatoric Uncertainty Control}
\paragraph{Multiple-Answer Examples Ratio Variation} 
We vary the ratio of WordNetMCQ1 and WordNetMCQ2 examples for WordNetMCQ2 queries. 
As shown in \Cref{tab:12variation}, Self-function vector yields the highest correlations across LLaMA2-7B, 13B, and 70B, suggesting that it more clearly reflects the effect of data ambiguity.
\begin{table}[H]
\centering
\resizebox{0.48\textwidth}{!}{%
\begin{tabular}{l|ccc}
\hline
Method & LLaMA2-7B & LLaMA2-13B & LLaMA2-70B \\
\hline
Total Entropy &  0.514 & 0.426 & 0.208 \\
Semantic Entropy   & -0.277 & -0.248 & -0.301 \\
UQ\_ICL            &  -0.093 & -0.097 & -0.380 \\
\cellcolor{gray!20}Function Vector   &  \cellcolor{gray!20}0.633 & \cellcolor{gray!20}0.429 & \cellcolor{gray!20}0.257 \\
\cellcolor{gray!20}Self-FV            & \cellcolor{gray!20}\textbf{0.640} & \cellcolor{gray!20}\textbf{0.435} & \cellcolor{gray!20}\textbf{0.292} \\
\hline
\end{tabular}
}
\caption{Spearman correlations ($\uparrow$) between WordNetMCQ1/2 ratio variation and uncertainty measures.}
\label{tab:12variation}
\end{table}

\paragraph{Label Noise Ratio Variation} 
We vary the flip ratio of example labels to control levels of AU. 
Results in \Cref{tab:label_noise_results} show that Self-function vector often attains higher correlations compared to other methods across LLaMA2 (7B/13B/70B) as well as Qwen2.5-7B and Mistral-7B, indicating that it captures the effect of label noise in a comparatively stable manner and that the mechanism generalizes across modern transformer-based LLMs.
This effect is particularly pronounced at larger model scales, consistent with the trend observed in \Cref{subsec:synthetic_data} that uncertainty control becomes more evident as model capacity increases.
\begin{table}[H]
\centering
\resizebox{0.49\textwidth}{!}{%
\begin{tabular}{l|l|ccccc}
\hline
Model & Method & WNMCQ1 & HellaSwag & GSM8K & AG News & Emotion  \\
\hline
\multirow{5}{*}{LLaMA2-7B} 
& Total Entropy & \textbf{0.177}& 0.061 & 0.143 &  0.739& \textbf{0.242} \\
& Semantic Entropy & 0.160 & 0.028 & \textbf{0.234} & 0.101 & 0.073  \\
& UQ\_ICL & 0.172 & 0.028 & 0.111 & 0.166 & 0.113  \\
& \cellcolor{gray!20}Function Vector & \cellcolor{gray!20}0.164 & \cellcolor{gray!20}0.057 & \cellcolor{gray!20}0.026 & \cellcolor{gray!20}0.668 & \cellcolor{gray!20}0.207 \\
& \cellcolor{gray!20}Self-FV            & \cellcolor{gray!20}0.167 & \cellcolor{gray!20}\textbf{0.064} & \cellcolor{gray!20}-0.023 & \cellcolor{gray!20}\textbf{0.746} & \cellcolor{gray!20}0.238 \\
\hline
\multirow{5}{*}{LLaMA2-13B}
& Total Entropy & 0.385 & 0.175 & 0.356 & 0.730 & \textbf{0.369}  \\
& Semantic Entropy & 0.364 & 0.105 & 0.015 & 0.323 & 0.159  \\
& UQ\_ICL & 0.422 & \textbf{0.200} & 0.161 & 0.443 & 0.211  \\
& \cellcolor{gray!20}Function Vector & \cellcolor{gray!20}0.316 & \cellcolor{gray!20}0.123 & \cellcolor{gray!20}0.360 & \cellcolor{gray!20}0.586 & \cellcolor{gray!20}0.364  \\
& \cellcolor{gray!20}Self-FV & \cellcolor{gray!20}\textbf{0.424} & \cellcolor{gray!20}0.183 & \cellcolor{gray!20}\textbf{0.368} & \cellcolor{gray!20}\textbf{0.741} & \cellcolor{gray!20}0.358  \\
\hline
\multirow{5}{*}{LLaMA2-70B}
& Total Entropy & 0.734 & 0.491 & 0.389 & \textbf{0.767} & 0.387 \\
& Semantic Entropy & 0.712 & 0.340 & 0.137 & 0.554 & 0.316 \\
& UQ\_ICL & 0.734 & 0.430 & 0.305 & \textbf{0.807} & 0.367 \\
& \cellcolor{gray!20}Function Vector & \cellcolor{gray!20}0.628 & \cellcolor{gray!20}0.406 & \cellcolor{gray!20}0.355 & \cellcolor{gray!20}0.751 & \cellcolor{gray!20}0.366 \\
& \cellcolor{gray!20}Self-FV & \cellcolor{gray!20}\textbf{0.798} & \cellcolor{gray!20}\textbf{0.566} & \cellcolor{gray!20}\textbf{0.476} & \cellcolor{gray!20}0.731 & \cellcolor{gray!20}\textbf{0.416} \\
\hline
\multirow{5}{*}{Qwen2.5-7B}
& Total Entropy & 0.737 & \textbf{0.580} & 0.499 & 0.705 & \textbf{0.517} \\
& Semantic Entropy & 0.605 & 0.145 & 0.085 & 0.262 & 0.298 \\
& UQ\_ICL & 0.690 & 0.456 & 0.023 & 0.306 & 0.181 \\
& \cellcolor{gray!20}Function Vector & \cellcolor{gray!20}0.742 & \cellcolor{gray!20}0.573 & \cellcolor{gray!20}0.527 & \cellcolor{gray!20}0.703 & \cellcolor{gray!20}0.500 \\
& \cellcolor{gray!20}Self-FV & \cellcolor{gray!20}\textbf{0.751} & \cellcolor{gray!20}0.574 & \cellcolor{gray!20}\textbf{0.542} & \cellcolor{gray!20}\textbf{0.714} & \cellcolor{gray!20}0.487 \\
\hline
\multirow{5}{*}{Mistral-7B}
& Total Entropy & 0.546 & 0.172 & 0.392 & \textbf{0.659} & \textbf{0.318} \\
& Semantic Entropy & 0.557 & -0.094 & 0.094 & 0.539 & 0.191 \\
& UQ\_ICL & 0.457 & 0.018 & 0.104 & 0.212 & 0.027 \\
& \cellcolor{gray!20}Function Vector & \cellcolor{gray!20}0.533 & \cellcolor{gray!20}0.170 & \cellcolor{gray!20}0.418 & \cellcolor{gray!20}0.653 & \cellcolor{gray!20}0.314 \\
& \cellcolor{gray!20}Self-FV & \cellcolor{gray!20}\textbf{0.657} & \cellcolor{gray!20}\textbf{0.173} & \cellcolor{gray!20}\textbf{0.436} & \cellcolor{gray!20}0.645 & \cellcolor{gray!20}0.311 \\
\hline
\end{tabular}
}
\caption{Spearman correlations ($\uparrow$) between label noise ratio variation and uncertainty.}
\label{tab:label_noise_results}
\end{table}


Together, these results show that mechanistic approaches exhibit stronger and more consistent correlations than alternative approaches, indicating that it more reliably reflects controlled sources of ambiguity.

\subsection{Epistemic Uncertainty Control}\label{subsec:eu_control}
\paragraph{OOD Query Variation}
This setup targets EU by moving queries progressively out of distribution. 
As shown in \Cref{tab:ood_query}, Function vector and Self-function vector show lower correlation with OOD variation compared to other entropy-based baselines across LLaMA2 (7B/13B/70B), Qwen2.5-7B, and Mistral-7B, indicating that they better isolate EU.

\begin{table}[H]
\centering
\resizebox{0.48\textwidth}{!}{%
\begin{tabular}{l|ccccc}
\hline
Method & LLaMA2-7B & LLaMA2-13B & LLaMA2-70B & Qwen2.5-7B & Mistral-7B \\
\hline
Total Entropy & 0.213 & 0.115 & 0.081 & 0.295 & -0.039 \\
Semantic Entropy   & 0.288 & 0.204 & 0.108 & \textbf{0.089} & 0.557 \\
UQ\_ICL            & 0.310 & 0.233 & 0.084 & 0.110 & 0.457 \\
\cellcolor{gray!20}Function Vector   & \cellcolor{gray!20}0.184 & \cellcolor{gray!20}0.089 & \cellcolor{gray!20}0.063 & \cellcolor{gray!20}0.294 & \cellcolor{gray!20}-0.0213 \\
\cellcolor{gray!20}Self-FV            & \cellcolor{gray!20}\textbf{0.148} & \cellcolor{gray!20}\textbf{0.076} & \cellcolor{gray!20}\textbf{0.026} & \cellcolor{gray!20}0.289 & \cellcolor{gray!20}\textbf{-0.0212} \\
\hline
\end{tabular}
}
\caption{Spearman correlations under OOD query variation, evaluated using WNMCQ1. Correlations closer to zero ($|\rho|\!\downarrow$) indicate better isolation of EU; bold marks the smallest magnitude per column.}
\label{tab:ood_query}
\end{table}

\begin{figure*}[!t]
    \centering
    \begin{subfigure}[t]{0.27\linewidth}
        \centering
        \includegraphics[width=\linewidth]{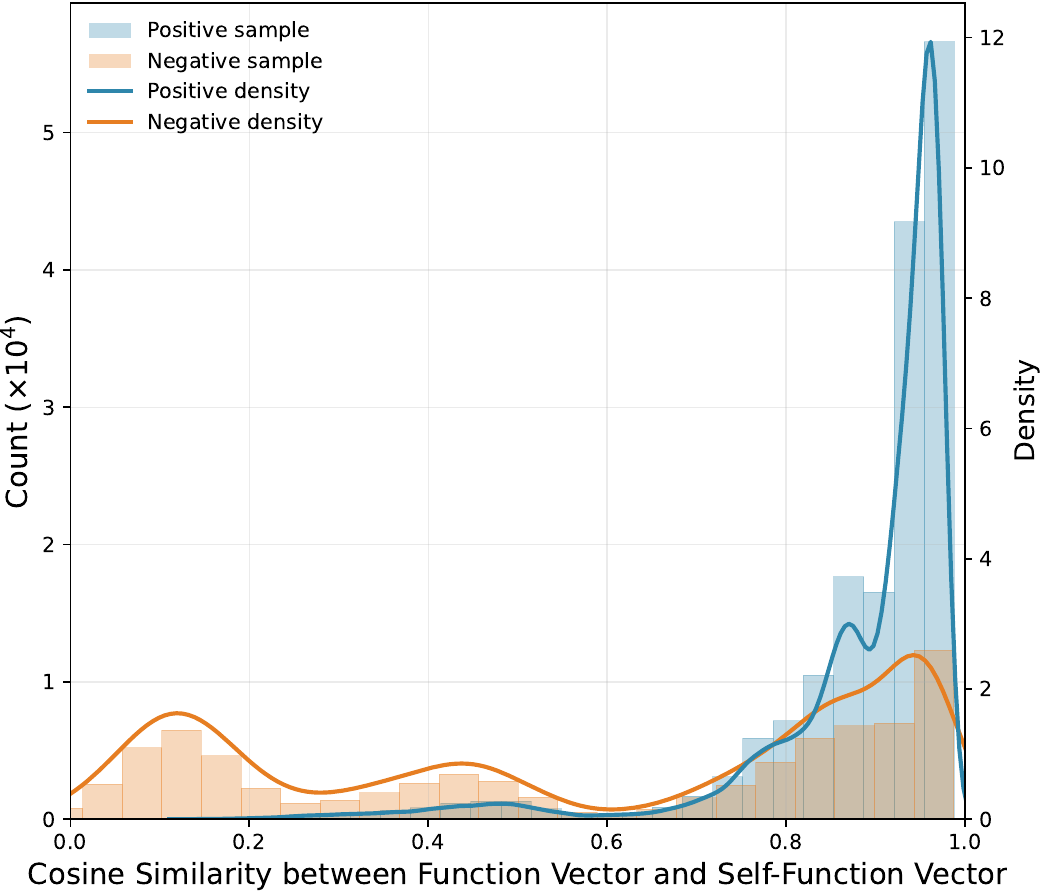}
        \caption{
        Cosine Similarity Distributions by Prediction Outcome}
        \label{fig:add_ablation1}
    \end{subfigure}
    \hfill
    \begin{subfigure}[t]{0.44\linewidth}
        \centering
        \includegraphics[width=\linewidth]{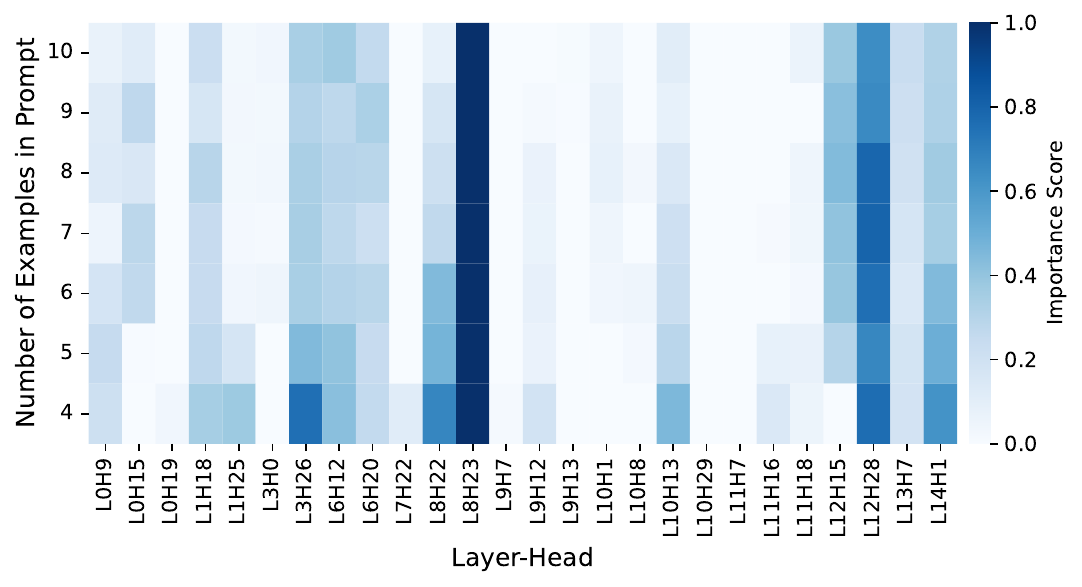}
        \caption{Top 20 Causal Heads}
        \label{fig:add_ablation2}
    \end{subfigure}
    \hfill
    \begin{subfigure}[t]{0.27\linewidth}
        \centering
        \includegraphics[width=\linewidth]{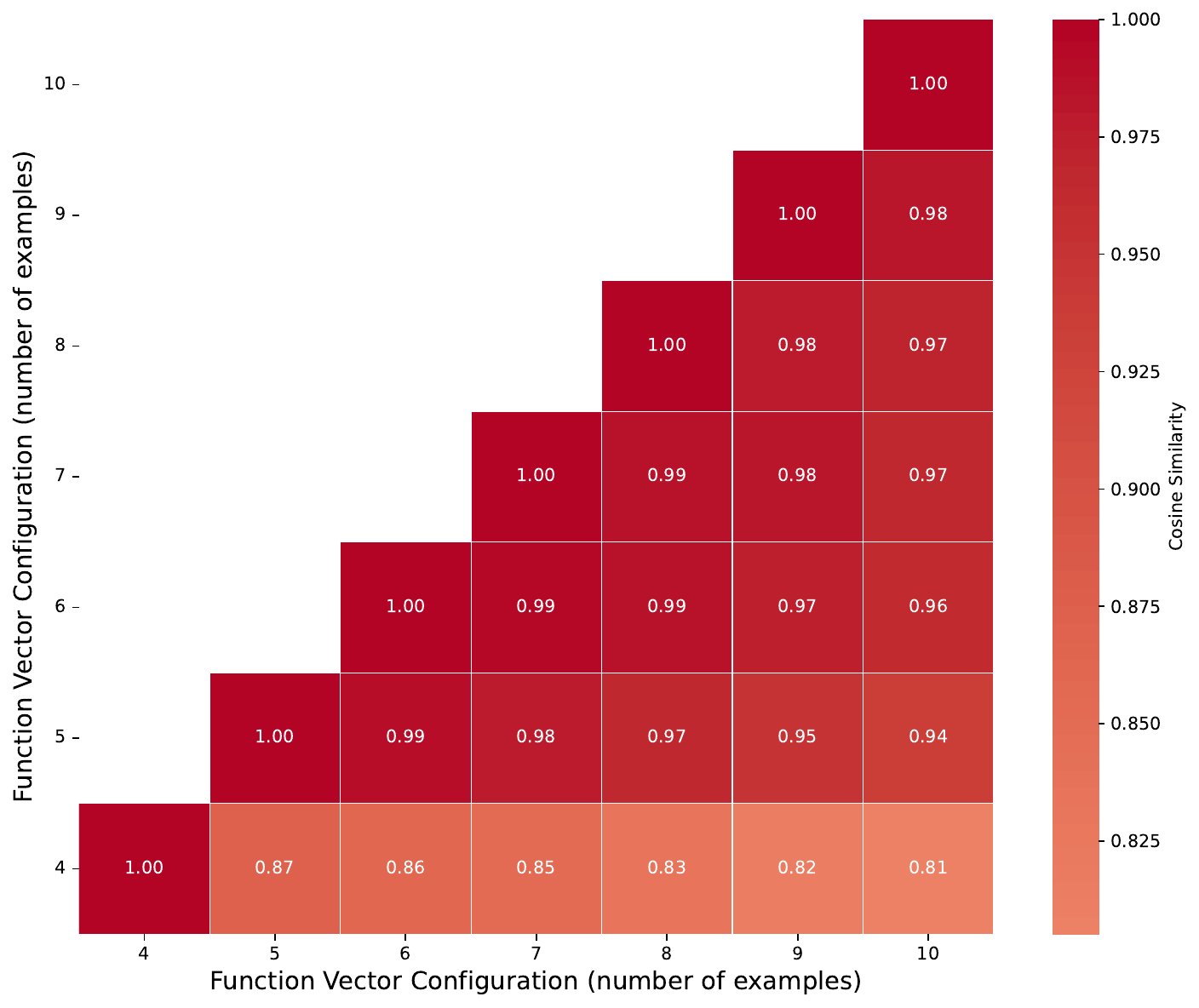}
        \caption{Function Vector Cosine Similarity Matrix}
        \label{fig:add_ablation3}
    \end{subfigure}
    \caption{Overall figure caption describing all three subplots. (a) This illustrates the distribution of cosine similarity between the self-function vectors and function vectors, separated by task correctness. Clear distinction between two distribution supports self-function vectors better reflect $P^\star$-specific internal task representation than function vectors. (b) This highlights the top 20 causal heads by their importance score. The x-axis indicates the number of examples used for CIE. The model identifies consistent causal heads across varying numbers of examples. (c) This shows cosine similarity between function vectors derived from various numbers of examples. The similarity of adjacent numbers converges to 1, indicating the consistent formation of the concept.}
    \label{fig:three_subplots}
\end{figure*}

\subsection{Hallucination Detection}
We also test on hallucination detection, a common downstream benchmark in uncertainty quantification. 
As shown in \Cref{tab:hallucination_detection}, mechanistic approaches generally outperform entropy-based baselines and remain competitive with or better than MaxProb; Lookback Lens, which relies on attention maps derived from the generation process, is less compatible with our single-token setting and underperforms accordingly.
Self-function vector performs comparably or better across several datasets, indicating practical benefits in trustworthy-related applications, with hyperparameters selected via a grid search (details in \Cref{app:experiment_details}).
\begin{table}[h]
\centering
\resizebox{0.48\textwidth}{!}{
\begin{tabular}{l|l|ccccc}
\hline
Model & Method & WNMCQ1 & Hellaswag & GSM8K & AG News & Emotion    \\
\hline
\multirow{5}{*}{LLaMA2-7B}
& Total Entropy & 0.872 & 0.608 & 0.554 & 0.843 & 0.646   \\
& Semantic Entropy & 0.859 & 0.578 & 0.626 & 0.532 & 0.538   \\
& UQ\_ICL & 0.884 & 0.575 & 0.657 & 0.805 & \textbf{0.681}  \\
& \cellcolor{gray!20}Function Vector & \cellcolor{gray!20}0.8989 & \cellcolor{gray!20}\textbf{0.624} & \cellcolor{gray!20}0.719 & \cellcolor{gray!20}\textbf{0.850} & \cellcolor{gray!20}0.651   \\
& \cellcolor{gray!20}Self-FV & \cellcolor{gray!20}\textbf{0.8993} & \cellcolor{gray!20}0.619 & \cellcolor{gray!20}\textbf{0.749} &  \cellcolor{gray!20}0.845 & \cellcolor{gray!20}0.647 \\
\hline
\multirow{5}{*}{LLaMA2-13B} 
& Total Entropy & 0.873 & 0.673 & 0.590 & 0.832 &   0.666 \\
& Semantic Entropy & 0.904 & 0.665 & 0.600 & 0.598 &   0.489 \\
& UQ\_ICL & \textbf{0.912} & 0.685 & 0.616 & \textbf{0.852} & \textbf{0.680}    \\
& \cellcolor{gray!20}Function Vector &  \cellcolor{gray!20}0.899 &  \cellcolor{gray!20}\textbf{0.696} &  \cellcolor{gray!20}0.627 & \cellcolor{gray!20}0.848 & \cellcolor{gray!20}0.676\\
& \cellcolor{gray!20}Self-FV & \cellcolor{gray!20}0.889 &  \cellcolor{gray!20}0.685 &  \cellcolor{gray!20}\textbf{0.638} &  \cellcolor{gray!20}0.845&  \cellcolor{gray!20}0.668 \\
\hline
\end{tabular}}
\caption{AUROC ($\uparrow$) on diverse text classification datasets. Mechanistic methods generally outperform entropy-based baselines.}
\label{tab:hallucination_detection}
\end{table}
PRR results \citep{fadeeva-etal-2024-fact} are provided in \Cref{app:prr} and show a consistent trend.
In summary, Self-function vector better reflects controlled AU/EU variations and performs competitively on conventional uncertainty quantification tasks, indicating that a mechanistic perspective offers a promising path for more reliable and interpretable ICL behavior.

\subsection{Ablation Study}\label{sec:ablation}
To validate our approach of AU measurement, we conduct a series of ablation studies, each targeting a component of the method.

\paragraph{Self-function vectors effectively capture prompt-specific concept representations}

We investigate whether self-function vectors more effectively capture the model’s internal representation of the target prompt $P^\star$, compared to function vectors which encode an average representation across prompts.

To test this, we separate prompts that the model generates the correct output with a clear concept from those it does not.
We then compare the cosine similarity between self-function vectors and function vectors across these two groups.

As shown in \Cref{fig:add_ablation1}, for the clear prompts that the model predicts the correct answer, cosine similarity is high, that is, the self-function aligns closely with the function vector. 
In contrast, the prompts that the model couldn't process the task show low cosine similarity, indicating the self-function vector diverges significantly from the function vector.

This trend is consistent across datasets, supporting that self-function vectors reflect the model's internal task representation specific to $P^\star$. The other experiments are reported in \Cref{subsec:ablation1}.

\paragraph{Further exploration}
We support that causal heads and corresponding function vectors indeed capture the internal representation of ICL through two experiments: Causal heads and function vectors under varying numbers of examples.

First, we observe a set of important heads that consistently emerge across different shot settings.
This stable selection implies the existence of certain heads that are crucial for the ICL task itself, rather than being specific to a particular ICL configuration, as shown in \Cref{fig:add_ablation2}.
Detailed results are provided in \Cref{subsec:ablation4}.

Second, the cosine similarity between the function vectors derived from prompts with adjacent numbers of in-context examples approaches 1 as the number of examples increases, as shown in \Cref{fig:add_ablation3}.
This convergence implies that function vectors indeed form a consistent conceptual understanding of the task. Additional results are presented in \Cref{subsec:ablation3}.
\section{Conclusion}
This work addresses the critical challenge of decomposing predictive uncertainty in ICL into its aleatoric and epistemic components. 
We introduce a novel methodology grounded in Bayesian views and mechanistic interpretability to directly quantify AU from the model's internal representations. 
Furthermore, we establish the first rigorous evaluation protocol specifically designed for uncertainty decomposition in ICL. 
Our experiments demonstrate not only a robust capacity to disentangle these two uncertainty sources, the also the practical utility in hallucination detection.
Ultimately, this research suggests a new direction that bridges mechanistic understanding with uncertainty estimation.
\section*{Limitations}
While this work seeks to bridge uncertainty quantification and mechanistic interpretability, several limitations remain.  

First, our use of function vectors as an approximation of posterior sampling over latent tasks is indirect and interpretive. 
The extent to which they capture true posterior structure remains uncertain, and warrants deeper theoretical analysis.  

Second, key hyperparameters, such as the intervention layer and the number of causal heads, require per-model tuning, which limits generalization across architectures.

We view these limitations as opportunities for future work, particularly in expanding task coverage, strengthening empirical validation, and refining the theoretical link between internal representations and Bayesian inference.

\section*{Acknowledgments}
This work was partly supported by the Institute of Information \& Communications Technology Planning \& Evaluation (IITP) grant funded by the Government of the Republic of Korea (Ministry of Science and ICT) (RS-2019-II191906, Artificial Intelligence Graduate School Program (POSTECH); RS-2022-II220959, (Part 2) Few-Shot Learning of Causal Inference in Vision and Language for Decision Making; RS-2023-00216011, Development of Artificial Complex Intelligence for Conceptually Understanding and Inferring like Humans); and by the High-Performance Computing Support Project, funded by the Government of the Republic of Korea (Ministry of Science and ICT) (RQT-25-070137).

\bibliography{custom}
\newpage
\appendix
\section{Synthetic Data Experiments} \label{sec:toy_experiments}

Here, we present a detailed description of the construction of the synthetic data in \Cref{subsec:synthetic_data}. We also report the corresponding results, which were omitted from the main paper due to space limitations.

\subsection{Construction of Synthetic Data}

We build binary classification ICL tasks on the Two Moons dataset. Following the synthetic ICL of \citet{jayasekera2025variationaluncertaintydecompositionincontext}, numerical values are rendered as string format for use in ICL prompts.

We construct binary classification tasks using the Two Moons dataset, where sampled examples are converted into string representations for ICL.

\texttt{make\_moons} in scikit-learn generates ICL examples 
\begin{itemize}
    \item $(\cos(x), \sin(x))$ for class 0,
    \item $(1-\cos(x), 0,5-\sin(x))$ for class 1, 
\end{itemize}
 with noise parameter which we set as $0.1$. Unless otherwise stated, we draw 30 examples.

\begin{figure*}[htbp]
    \centering
    \begin{subfigure}{0.32\textwidth}
        \centering
        \includegraphics[width=\linewidth]{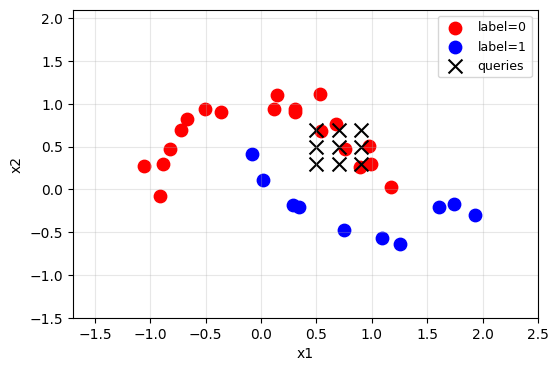}
        \caption{default}
    \end{subfigure}
    \hfill
    \begin{subfigure}{0.32\textwidth}
        \centering
        \includegraphics[width=\linewidth]{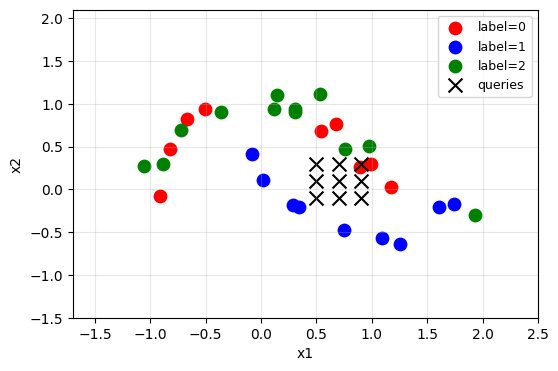}
        \caption{multiple-answer examples}
    \end{subfigure}
    \hfill
    \begin{subfigure}{0.32\textwidth}
        \centering
        \includegraphics[width=\linewidth]{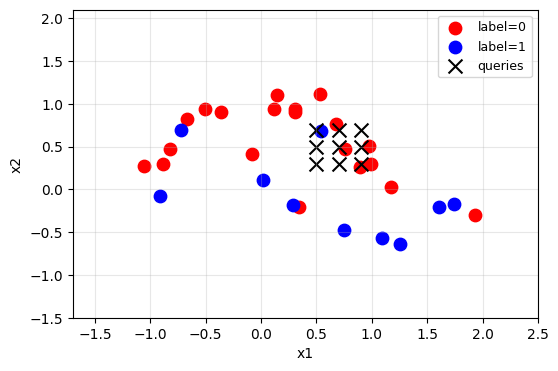}
        \caption{label flip}
    \end{subfigure}
    
    \vskip\baselineskip
    \begin{subfigure}{0.32\textwidth}
        \centering
        \includegraphics[width=\linewidth]{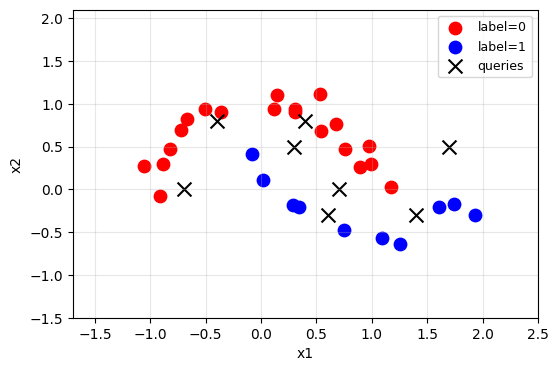}
        \caption{ambiguous query}
    \end{subfigure}
    \hfill
    \begin{subfigure}{0.32\textwidth}
        \centering
        \includegraphics[width=\linewidth]{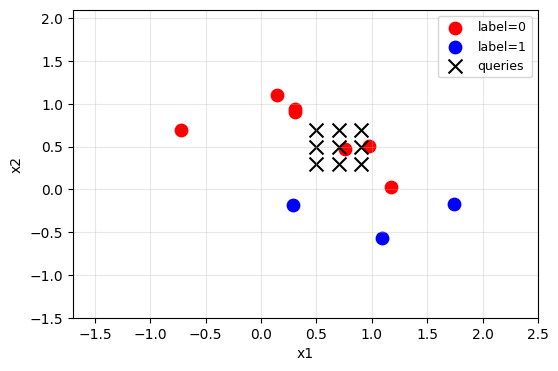}
        \caption{example number}
    \end{subfigure}
    \hfill
    \begin{subfigure}{0.32\textwidth}
        \centering
        \includegraphics[width=\linewidth]{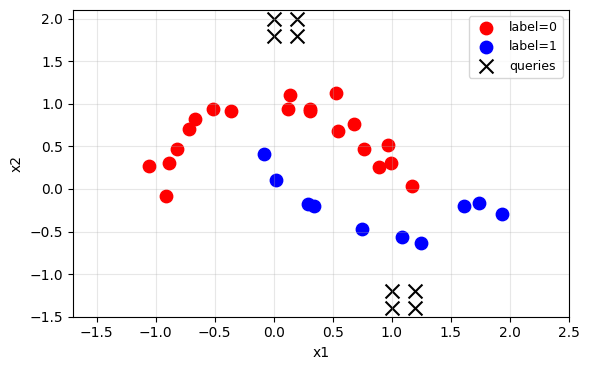}
        \caption{OOD query}
    \end{subfigure}

    \caption{Visualization of synthetic data for each task}
    \label{fig:synthetic_data_distr}
\end{figure*}

To examine the behavior of AU and EU, we perturb both examples and queries under five families of tasks---Multiple-answer examples, Label flip, Ambiguous query, Example number and OOD query. The implementation of each task is summarized in \Cref{tab:sanity_check_implementation} and \Cref{fig:synthetic_data_distr}.

\paragraph{Multiple-answer Examples}
From the class posteriors evaluated at each example location, we compute the probability margin between the classes. We then relabel the bottom–\(k\%\) of examples by this margin as multi-label, which we denoted as $2$. We vary the ratio of multi–labeled examples to single–labeled from \(0\%\) to \(100\%\) (default).
Queries are selected near the decision boundary to simulate the ambiguity that arises when, given single-labeled context, the model must choose one answer from two valid options.

\paragraph{Label Flip}
A fraction of example labels is flipped uniformly at random. 
The flip ratio is swept from \(0\%\) (default) to \(50\%\); ratios above \(50\%\) are symmetric and thus omitted.

\paragraph{Ambiguous Query}
To move queries toward the class boundary, we contract their radius with a nonnegative degree parameter \(d\):
\begin{itemize}
    \item $\Bigl(\tfrac{\cos x}{1+d},\; \tfrac{\sin x}{1+d/3}\Bigr)$,
    \item $\Bigl(1-\tfrac{\cos x}{1+d},\; 0.5-\tfrac{\sin x}{1+d/3}\Bigr)$.
\end{itemize}
We sweep \(d\) from \(0\) to \(100\).

\paragraph{Example Number}
We vary the number of examples from \(0\) to \(50\).

\paragraph{OOD Query}
To place queries Out-Of-Distribution (OOD), we translate small grids away from the moons:
{\small
\begin{itemize}
    \item $\{(0+i,1+j+0.2\times \text{degree}):i,j\in\{0,0.2\}\}$,
    \item $\{(1+i,-0.6+j-0.2\times \text{degree}):i,j\in\{0,0.2\}\}$.
\end{itemize}
}
We sweep \(d\) from \(0\) to \(11\).

Among these, we retain only tasks in which either AU or EU consistently increase while the other remains stable, thereby composing the WordNet evaluation protocol.

\begin{table*}[t]
    \centering
    \begin{tabularx}{\textwidth}{|l|X|}
        \hline
        \textbf{Tasks} & \textbf{Control} \\ \hline
Multiple-answer examples & For examples with similar class-probability logits, map their labels to $2$. \\ \hline
Label flip & Randomly flip example labels. \\ \hline
Ambiguous query & Move the query closer to the decision boundary of the Two Moons dataset. \\ \hline
Example number & Change the number of in-context examples. \\ \hline
OOD query & Move the query further from the decision boundary of the Two Moons dataset. \\ \hline
    \end{tabularx}
    \caption{Task Implementation on Synthetic Data}
    \label{tab:sanity_check_implementation}
\end{table*}

\subsection{Benchmark Validation Results}

\begin{figure*}[htbp]
    \centering
    \begin{subfigure}{0.77\textwidth}
        \centering
        \includegraphics[width=\linewidth]{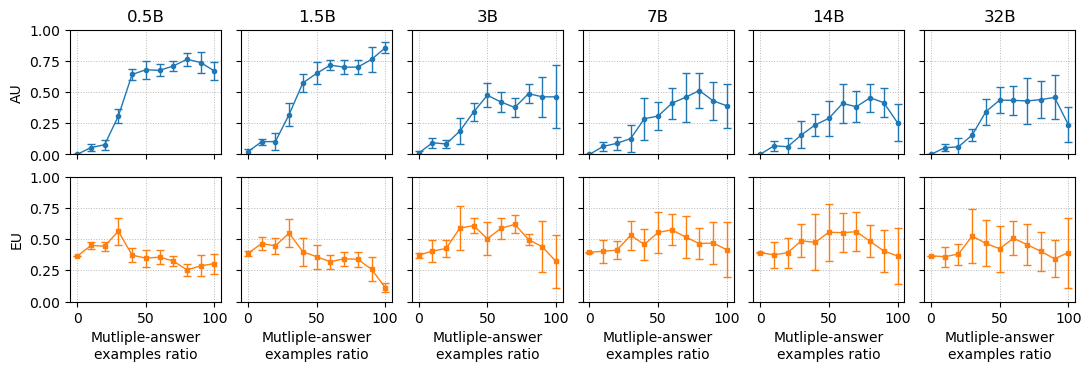}
        \caption{multi-answered examples}
    \end{subfigure}

    \begin{subfigure}{0.77\textwidth}
        \centering
        \includegraphics[width=\linewidth]{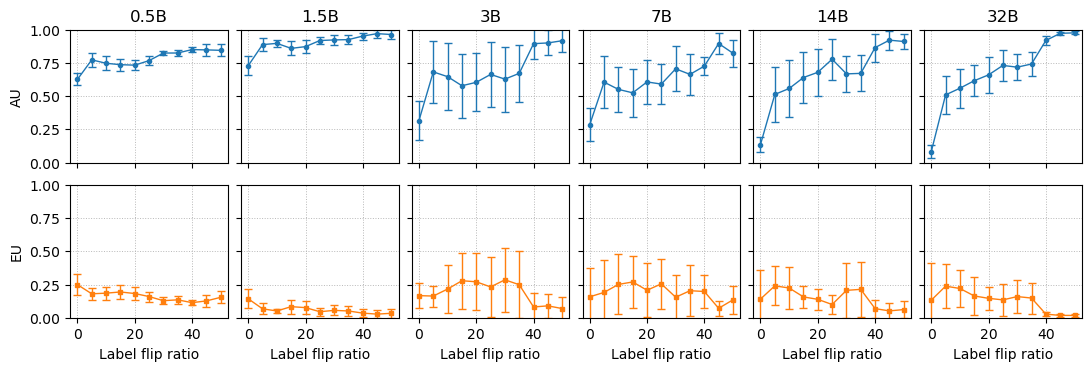}
        \caption{label flip}
    \end{subfigure}

    \begin{subfigure}{0.77\textwidth}
        \centering
        \includegraphics[width=\linewidth]{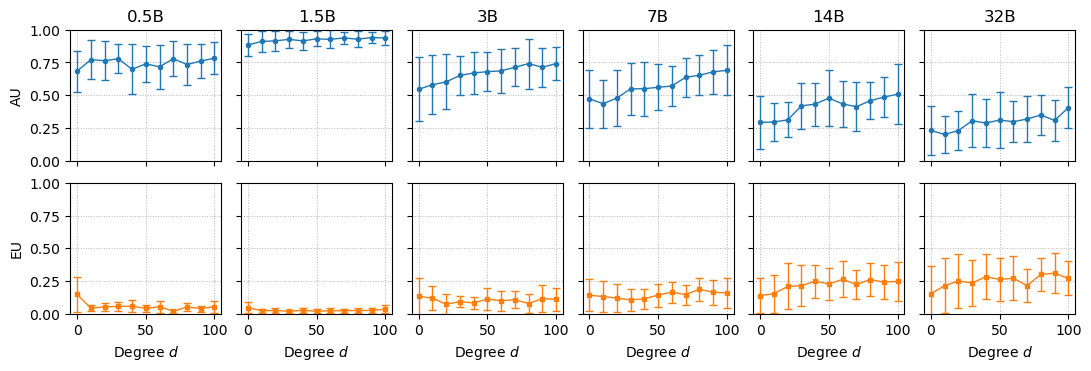}
        \caption{ambiguous query}
    \end{subfigure}

    \begin{subfigure}{0.77\textwidth}
        \centering
        \includegraphics[width=\linewidth]{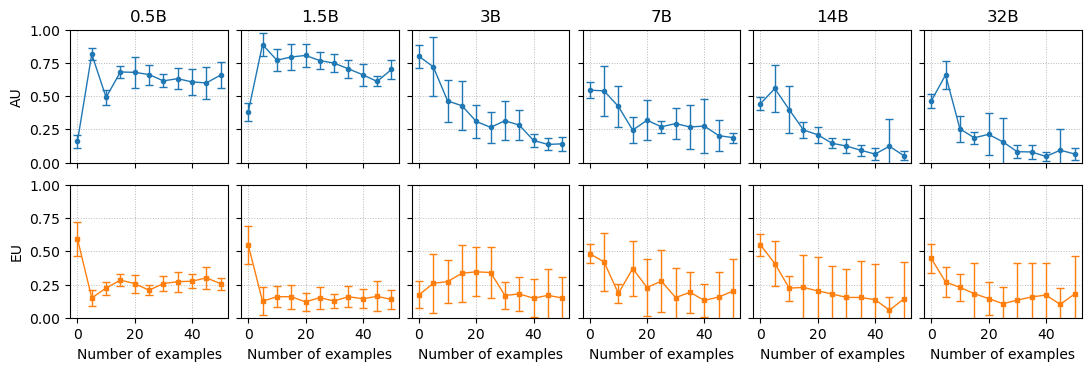}
        \caption{example number}
    \end{subfigure}

    \begin{subfigure}{0.77\textwidth}
        \centering
        \includegraphics[width=\linewidth]{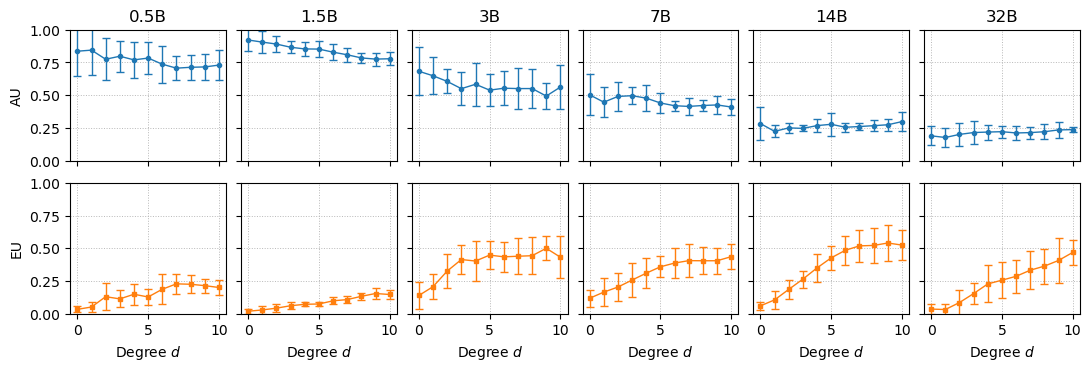}
        \caption{OOD query}
    \end{subfigure}

    \caption{AU / EU change under each perturbation.}
    \label{fig:benchmark_validation}
\end{figure*}

We apply the uncertainty decomposition method proposed by \citet{jayasekera2025variationaluncertaintydecompositionincontext} to the Qwen2.5 models~\citep{qwen2025qwen25technicalreport}, spanning from 0.5B to 32B parameters. \Cref{fig:benchmark_validation} illustrates the behaviors of AU and EU across different model scales under various perturbation settings.

Across all tasks, we observe distinct AU and EU fluctuations depending on model size. We claim that effective control of AU and EU requires a model sufficiently large to perform ICL reliably.

For larger models (typically above 7B parameters), the Multi-Answered Examples task exhibits an increase in AU, while EU fluctuates without a clear trend. The observed drop in AU at the $100\%$ ratio setting likely results from a reduced number of valid answer candidates, as this configuration excludes multi-labeled examples.

Similarly, in the Label Flip task, AU consistently increases with only minor EU fluctuations. Consequently, we adopt the Multi-Answered Examples and Label Flip tasks as benchmark settings for AU control.

In contrast, the Ambiguous Query task demonstrates simultaneous increases in both AU and EU, while the Example Number task shows decreases in both measures. Therefore, we exclude these tasks from the evaluation protocol.

Finally, the Out-of-Distribution (OOD) Query task yields stable EU and increasing AU as model size grows, making it a suitable protocol for EU control.

\section{Evalution Metric Comparison}

In this section, we investigate several candidate metrics to identify the most appropriate one for our task.

\subsection{Candidate Metrics}

We first provide a brief description of each metric under consideration, outlining their key characteristics and differences.

\paragraph{Pearson’s correlation.}
Pearson’s correlation quantifies the linear association between two continuous variables. It is defined as the standardized covariance and ranges from $-1$ to $+1$, with extremal values indicating perfect positive or negative linear dependence. Under joint normality, it fully characterizes dependence, making uncorrelatedness equivalent to independence.

\paragraph{Spearman’s rank-correlation.}
Spearman’s correlation is a nonparametric measure of monotonic association. It is obtained by replacing observed values with ranks and then computing the Pearson correlation on these ranks. By focusing on order rather than magnitude, it is robust to non-normality, nonlinear transformations, and outliers.

\paragraph{Goodman-Kruskal’s gamma.}
Gamma is a measure of monotonic association between two ordinal variables. It is defined as the difference between concordant and discordant pairs, normalized by their sum. Because ties are excluded, gamma is most suitable when relative ordering is of primary interest rather than their absolute differences.

\paragraph{Kendall’s Tau-b.}
Tau-b also measures ordinal association but explicitly corrects for ties, providing a more balanced statistic when tied observations are common.

\paragraph{Kendall’s Tau-c.}
While conceptually similar to Kendall’s tau-b, Tau-c introduces an adjustment for table dimensionality, ensuring values remain between $-1$ and $+1$ even when the two variables have different category counts.

\paragraph{Somers’ d.}
Somers’ d is an asymmetric extension of Kendall’s tau, treating one variable as independent and the other as dependent, and correcting only for ties on the independent variable.
\subsection{\texorpdfstring{Experiment Details of \Cref{subsec:evaluation_metric}}{Experiment Details of section 4.3}}
\label{subsec:metric_details}

We present a toy example in which we empirically compare the performance of these metrics. Based on this analysis, we determine the most suitable metric to employ in our evaluation.

Recall that each task has a uniform number of data across its ordinal level, while measured uncertainty is on the real line. To find the most suitable metric that captures the correlation between levels and uncertainties, we first generated two random variables with correlation $\rho$ as follows:

\begin{itemize}
    \item $X = \frac{\sqrt{1+\rho}+\sqrt{1-\rho}}{2} X_1 +  \frac{\sqrt{1+\rho}-\sqrt{1-\rho}}{2} X_2$,
    \item $X = \frac{\sqrt{1-\rho}+\sqrt{1-\rho}}{2} X_1 +  \frac{\sqrt{1+\rho}+\sqrt{1-\rho}}{2} X_2$,
\end{itemize}
where $X_1$ and $X_2$ are sampled from a normal distribution. The variable $X$ is discretized into ordered categories with approximately uniform frequencies via quantile-based binning, while $Y$ remains continuous and is transformed by monotonic functions such as exponential, arctan, cubic, and cubic-root to induce nonlinear dependence.

\begin{figure}[htbp]
    \centering
    \includegraphics[width=0.95\linewidth]{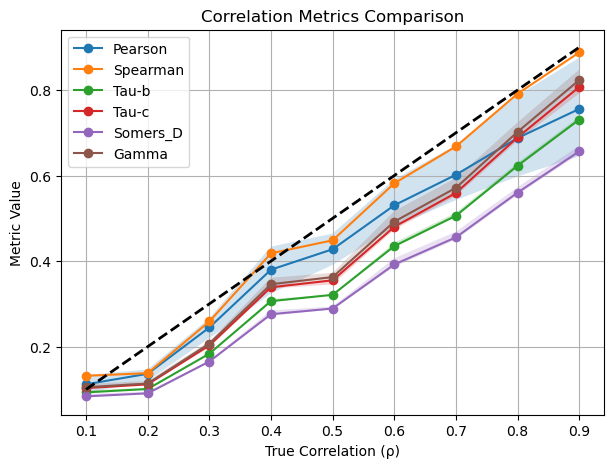}
    \caption{Metric evaluation}
    \label{fig:metric_evaluation}
\end{figure}

This procedure yielded a non-symmetric contingency structure, reflecting an ordered categorical independent variable and a continuous dependent variable with a non-linear dependence. We then measure every candidate metrics for various $\rho$. The result is in \Cref{fig:metric_evaluation}. We observe that Spearman’s rank correlation most faithfully captures the underlying correlation with the lowest variance, and is therefore adopted as the evaluation metric for our benchmark.

\section{Prompt Examples}

\noindent
In \Cref{subsec:eu_control}, we manipulated the OOD (Out-of-Distribution) levels of the query.
Below, we present examples of the prompts used at each OOD level.
As the OOD level increases, the query text becomes progressively noisier, while the core semantic meaning remains the same.

\begin{itemize}
  \item \textbf{Level 0 Example (Original)}
  \begin{verbatim}
Question: Which of the following is/are 
types of debris?
Choices:
A. leg of lamb
B. sexton
C. shopper
D. slack
Answer: D
  \end{verbatim}

  \item \textbf{Level 1 Example}
  \begin{verbatim}
Question: Among the options, which one(s) 
classify as types of debris?
Choices:
A. leg of lamb
B. sexton
C. shopper
D. slack
Answer: D
  \end{verbatim}

  \item \textbf{Level 2 Example}
  \begin{verbatim}
Answer: D 

Answer: Question: Among the options, 
which !one(s) classify as 
types of de@bris?
Choices:
A. leg of lamb
B. sexton
C. shopper
D. slack
Answer: D
  \end{verbatim}

  \item \textbf{Level 3 Example}
  \begin{verbatim}
Question: Among the options, which 
$one(s) classify as types# 
of de!br$is?
Choices:
A. leg of lamb
B. sexton
C. shopper
D. slack
Answer: D
  \end{verbatim}
\end{itemize}
\section{Experiment Details}\label{app:experiment_details}
\paragraph{Baseline}We consider Total Entropy, Semantic Entropy \citep{kuhn2023semantic}, and UQ\_ICL \citep{ling-etal-2024-uncertainty} as representative baselines for uncertainty quantification and decomposition. Total Entropy measures predictive uncertainty directly from the token-level probability distribution. Semantic Entropy computes entropy over semantically clustered generations, thereby discounting superficial lexical variation. UQ\_ICL decomposes total uncertainty in in-context learning into aleatoric and epistemic components, attributing the former to inherent data ambiguity and the latter to variability arising from model parameters or prompt configurations.

In addition, \citet{wang2025on} recently introduced a principled framework for subjective uncertainty quantification and calibration in natural language generation, including discussions relevant to in-context learning, while we respectfully acknowledge the contribution of \citet{jayasekera2025variationaluncertaintydecompositionincontext}, who proposed a variational method for decomposing uncertainty in in-context learning. However, these approaches are excluded from our direct comparisons, as they typically require external APIs or well instruction-tuned models to generate auxiliary data, or rely on explicitly varying the number of in-context examples.

\paragraph{Data} We use datasets from HuggingFace Datasets \citep{hf_datasets}: AGNews \citep{ag_news_citation}, Emotion \citep{emotion_citation}, HellaSwag \citep{hellaswag_citation}, and GSM8K \citep{gsm8k_citation}, where AGNews and Emotion are evaluated in the standard ICL setting with open-ended responses \citep{liu2025IterativeVectors, ling-etal-2024-uncertainty}, and HellaSwag and GSM8K are formatted as multiple-choice questions \citep{ye2024benchmarking}, consistent with our WNMCQ1 setup. For datasets with pre-defined train/test splits, we use them directly; otherwise, we randomly split into train/test with an 8:2 ratio. The validation set used for causal effect estimation is drawn from 10\% of the training data. The number of in-context examples is fixed to 10, both for causal effect estimation and for inference. In aleatoric uncertainty control experiments, the proportion of WordNetMCQ1 examples is varied over \{0, 30, 60, 90, 100\}\%, and the label-flipping ratio for noise injection follows the same schedule.

\paragraph{Hyperparameters} By default (all tasks except hallucination detection), we intervene at one-third of the transformer: layer 10 for LLaMA2-7B and layer 13 for LLaMA2-13B, using the top-20 causal heads. For hallucination detection, we select hyperparameters via a small grid search with \texttt{interv\_layer} $\in \{7,8,9,10,11,12,13,14,15\}$ and \texttt{num\_top\_heads} $\in \{8,10,12,14,16,18,20\}$, reporting the best validation configuration.
\section{Additional Ablation Results}\label{app:ablation}

Here, we provide experiment settings and results on every dataset that are omitted in \Cref{sec:ablation}.

\subsection{Self-function vectors effectively capture prompt-specific concept representations}\label{subsec:ablation1}
\begin{figure}[htbp]
    \centering
    \includegraphics[width=\linewidth]{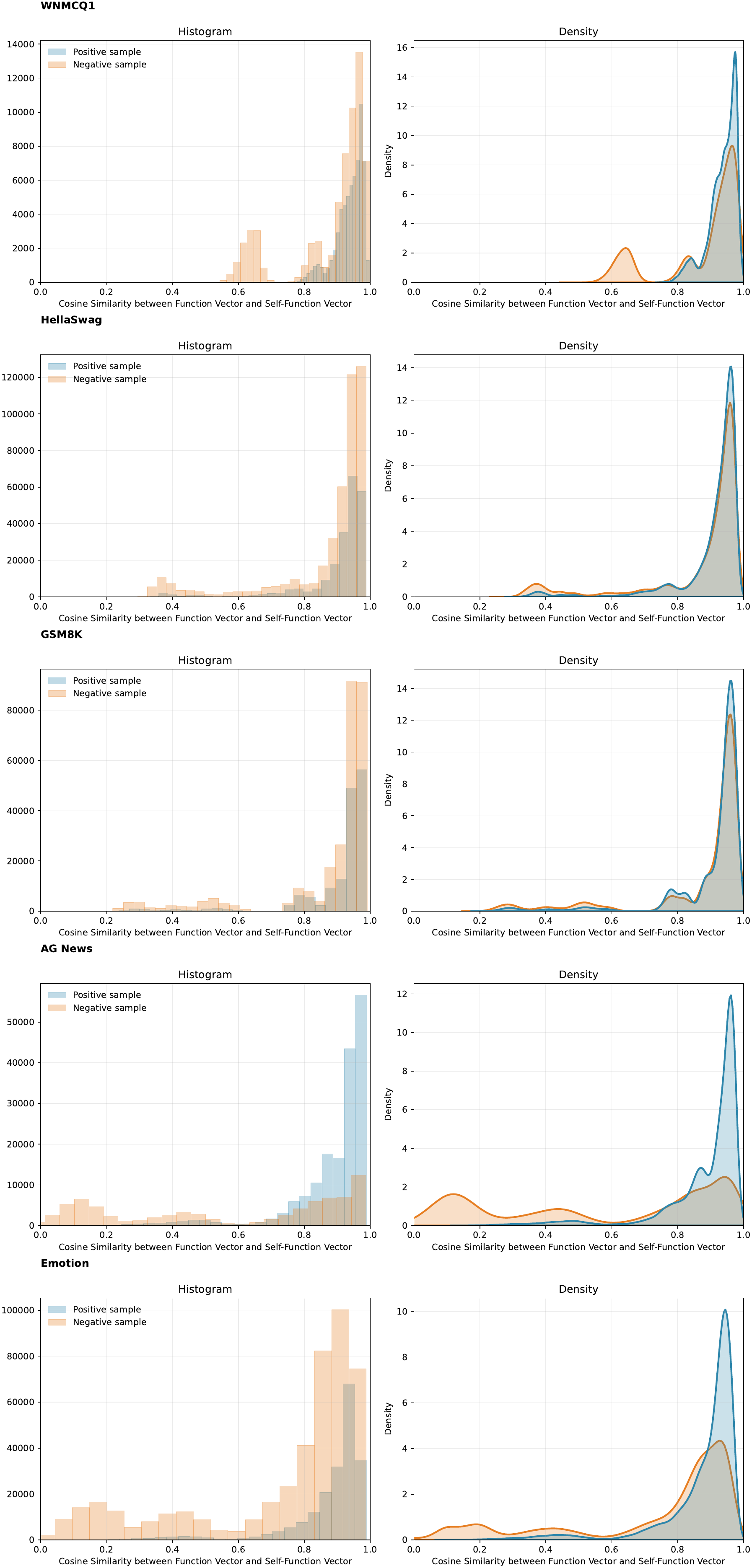}
    \caption{Cosine similarity distributions between self-function vectors and original task function vectors across different numbers of in-context examples $[0, 1, \cdots, 10]$. Incorrect samples in AG News and Emotion exhibit lower similarity, while multiple-choice datasets (WNMCQ1, Hellaswag, GSM8K) show more stable distributions across prompt configurations.}
    \label{fig:hist_density_grid}
\end{figure}

Here, we present results on every dataset that are omitted in the main section.

We analyzed the relationship between the number of examples in the prompt $[0, 1, \cdots, 10]$ and the similarity between the task-specific head’s self-function vector and the original task’s function vector. 
For each prompt configuration, we visualized the cosine similarity distributions using histograms and kernel density estimation (KDE), conditioned on whether the model’s prediction (with or without self-function vector intervention) was correct or incorrect.
The results indicate that, for the AG News and Emotion datasets, incorrect samples tend to exhibit lower cosine similarity between the self-function vector and the original task function vector. 
In contrast, datasets such as WNMCQ1, Hellaswag, and GSM8K—which adopt a multiple-choice question format—showed more consistent similarity distributions, likely because their conceptual representations remain stable across examples.
\subsection{Single Top-\texorpdfstring{$k$}{k} Intervention vs.\ Ensemble Variants}\label{subsec:ensemble}

We compare our top-$k$ single intervention against ensemble-style variants that aggregate AU estimates over $N$ randomly sampled interventions from $\mathcal{S}_T$, in order to empirically justify the $N=1$ design choice described in \Cref{method:quantification}.

Concretely, for each prompt we compute the AU estimate using the top-$k$ self-function vector (our default) and compare it against the mean AU estimate obtained from $N \in \{1, 3, 5, 10, 20, 30, 40, 50\}$ randomly sampled interventions. We report the mean absolute difference, its 95\% confidence interval, and relative variance across prompts on WNMCQ1 with LLaMA2-7B.

\begin{table}[H]
\centering
\resizebox{0.48\textwidth}{!}{%
\begin{tabular}{c|cccc}
\hline
$N$ & Mean Diff & 95\% CI & Variance & Rel.\ Variance \\
\hline
1  & 0.0420 & [0.0257, 0.0583] & 0.00223 & 0.00156 \\
3  & 0.0331 & [0.0183, 0.0478] & 0.00148 & 0.00100 \\
5  & 0.0310 & [0.0178, 0.0443] & 0.00127 & 0.00085 \\
10 & 0.0272 & [0.0152, 0.0393] & 0.00100 & 0.00067 \\
20 & 0.0260 & [0.0145, 0.0375] & 0.00091 & 0.00061 \\
30 & 0.0244 & [0.0141, 0.0348] & 0.00079 & 0.00053 \\
40 & 0.0240 & [0.0141, 0.0340] & 0.00075 & 0.00050 \\
50 & 0.0244 & [0.0145, 0.0343] & 0.00077 & 0.00051 \\
\hline
\end{tabular}}
\caption{Mean absolute difference between the top-$k$ single intervention AU estimate and a random ensemble of $N$ members (average AU $= 1.302$). Differences decrease rapidly and plateau after $N\approx 10$, remaining below 2.1\% of the average AU.}
\label{tab:ensemble_comparison}
\end{table}

As shown in \Cref{tab:ensemble_comparison}, the mean absolute difference between the top-$k$ estimate and the ensemble converges quickly: it drops from 0.042 at $N=1$ to roughly 0.024 by $N=10$ and remains stable thereafter. Relative to the average AU of 1.302, the gap at $N=1$ is approximately 3.2\%, and converges to $\sim$1.8\% as $N \to 50$. These marginal differences confirm that the top-$k$ single intervention already captures the dominant effect, with additional ensembling providing limited practical gain relative to the added computational cost.
\subsection{Impact of the number of shots on causal head selection}\label{subsec:ablation4}
\begin{figure*}[t]
    \centering
    \includegraphics[width=\linewidth]{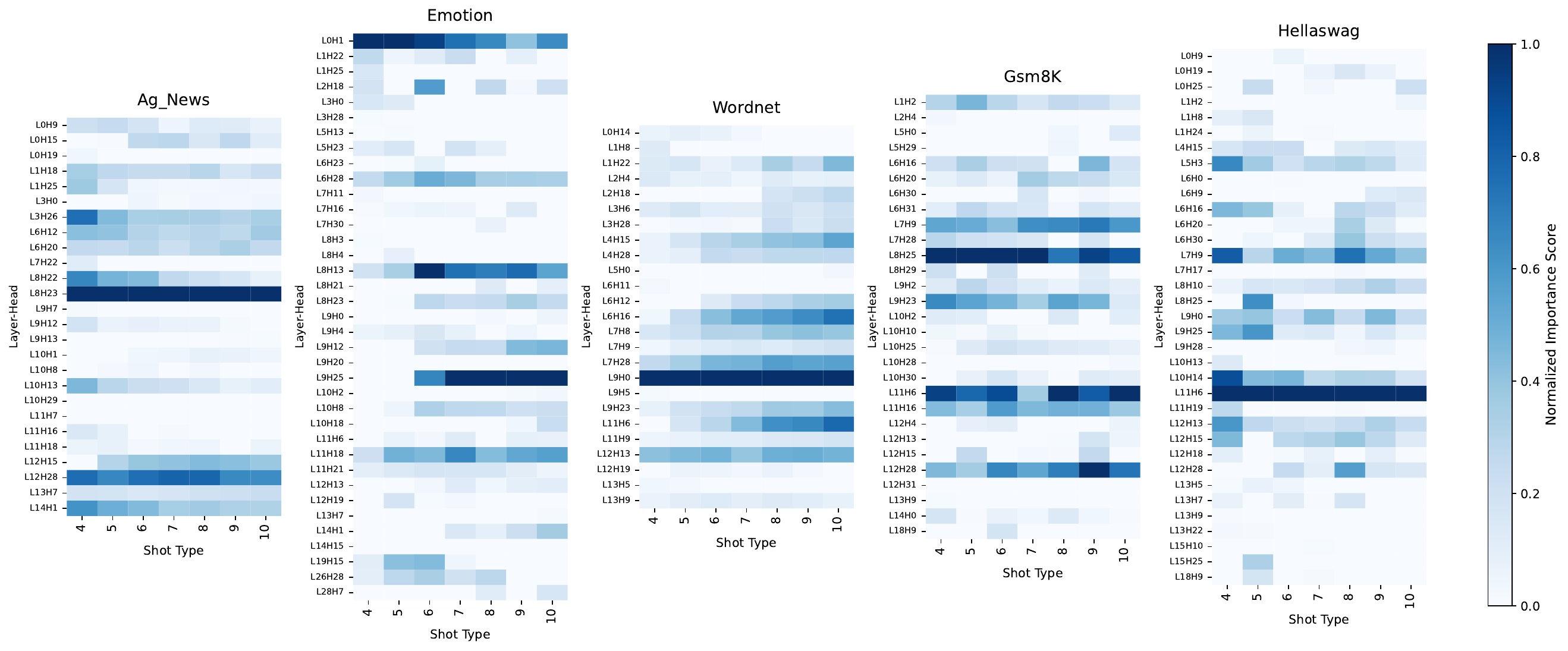}
    \caption{Consistent task-specific causal heads across different shot configurations, suggesting their activations can be effectively leveraged.}
    \label{fig:top_heads_normalized_heatmap_all}
\end{figure*}

In exploring the identification of causal heads, we varied the number of examples in the prompt to examine how robustly task-specific heads can be discovered and how effectively their activations can be leveraged. Specifically, we varied the number of examples from 4 to 10 and identified causal heads for each task. Our analysis revealed that, regardless of the configuration, there consistently exist task-specific heads that play a crucial role, which further supports the effectiveness of our proposed method.
\subsection{Function vector converges as number of shots increases}\label{subsec:ablation3}
\begin{figure}[htbp]
    \centering
    \includegraphics[width=0.9\linewidth]{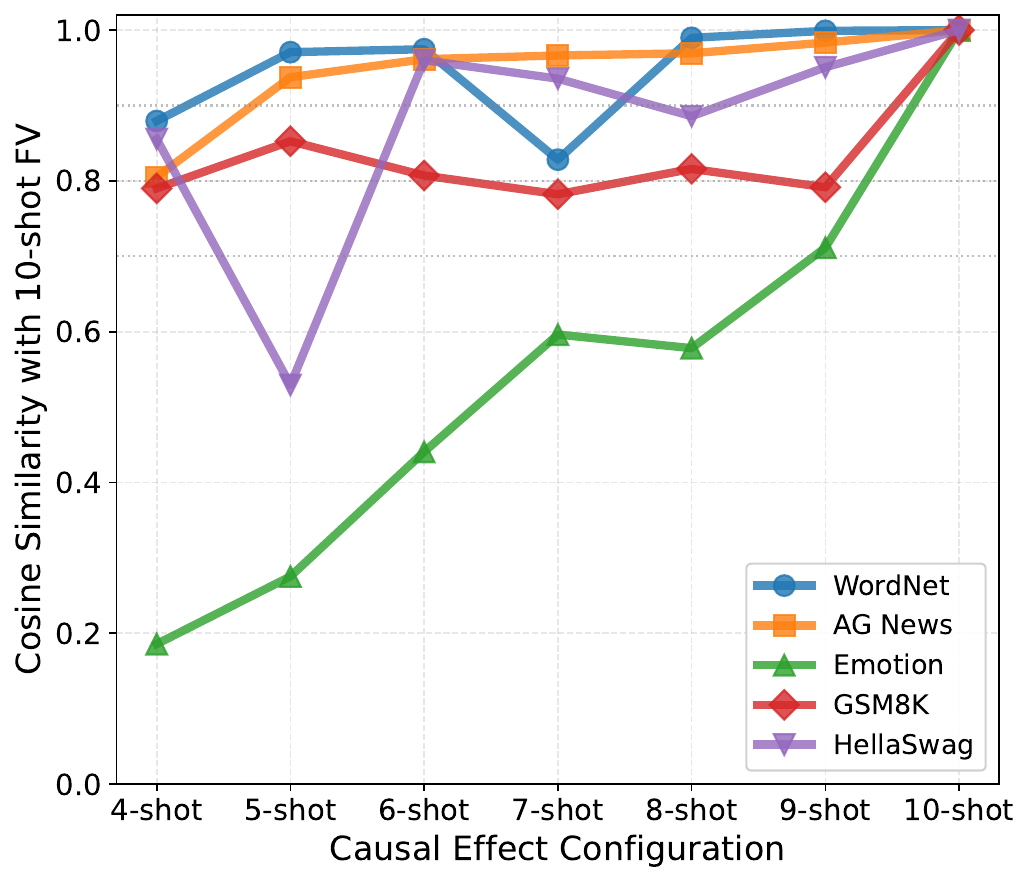}
    \caption{Cosine similarity between function vectors across tasks as the number of in-context examples increases from 4 to 10.}
    \label{fig:fv_simialrity}
\end{figure}
Here, we provide experiment settings and results on every dataset that are omitted in the main section.

To further investigate the stability of function vectors derived under different numbers of in-context examples, we analyzed their behavior in the causal head selection setting described in Section \Cref{subsec:ablation4}.
For each given task T, we obtained multiple function vectors that represent the task semantics under varying prompt configurations. Specifically, we used the LLaMA-2-7B model and identified causal heads for each task while varying the number of in-context examples from 4 to 10.

We then compared the resulting function vectors across adjacent configurations (e.g., between 4 and 5 examples, 5 and 6 examples, etc.) by computing their cosine similarities. We observed that, as the number of examples increases, the cosine similarity between function vectors derived from neighboring configurations approaches 1, indicating that the function representation of the task converges to a stable direction in representation space. This suggests that, once sufficient contextual information is provided, the model forms a consistent and robust conceptual understanding of the given task, rather than being sensitive to minor variations in the number of examples.
\section{Additional Baselines and Related Work}\label{app:additional_baselines}

\subsection{Related Work: RAUQ}\label{app:rauq_discussion}
\citet{vazhentsev2025uncertaintyaware} propose RAUQ, an unsupervised UQ method that leverages attention-head signals within transformer layers to estimate prediction confidence in standard text generation. While RAUQ shares the broad motivation of grounding uncertainty in internal model components, it targets total UQ in text generation rather than aleatoric--epistemic decomposition in ICL, making it complementary to rather than directly comparable with our work.

\subsection{MaxProb and Lookback Lens}\label{app:maxprob_lookback}
We extend the baseline comparison to two additional methods: MaxProb \citep{hendrycks2017baseline}, a canonical confidence baseline using the maximum softmax probability, and Lookback Lens \citep{chuang-etal-2024-lookback}, a recent attention-based hallucination detection method.

\paragraph{Uncertainty Decomposition Protocol} \Cref{tab:additional_baselines} reports Spearman correlations of MaxProb and Lookback Lens under our AU- and EU-control settings on LLaMA2-7B. Both methods yield small or inconsistent correlations across the AU-control settings and noticeably larger OOD correlations than Self-FV, reflecting that they are designed for overall confidence estimation rather than disentangling AU from EU. These results further motivate specialized decomposition methods for ICL.

\begin{table}[H]
\centering
\resizebox{0.49\textwidth}{!}{%
\begin{tabular}{l|l|cccccc|c}
\hline
Model & Method & Multi-Ans. & WNMCQ1 & HellaSwag & GSM8K & AG News & Emotion & OOD \\
\hline
\multirow{2}{*}{LLaMA2-7B}
& MaxProb       & 0.249 & 0.167 & 0.022 & 0.220 & 0.145 & 0.097 & 0.153 \\
& Lookback Lens & -0.970 & 0.0006 & 0.0001 & 0.322 & 0.188 & -0.035 & 0.406 \\
\hline
\end{tabular}
}
\caption{MaxProb and Lookback Lens on the AU- (Multi-Ans., WNMCQ1, HellaSwag, GSM8K, AG News, Emotion) and EU- (OOD) control evaluation protocol for LLaMA2-7B, complementing the main-text results in \Cref{tab:12variation,tab:label_noise_results,tab:ood_query}. AU columns ($\uparrow$); OOD ($|\rho|\!\downarrow$).}
\label{tab:additional_baselines}
\end{table}

\paragraph{Hallucination Detection} \Cref{tab:hallucination_detection_additional} reports AUROC for MaxProb and Lookback Lens on LLaMA2-7B, complementing the main hallucination detection results in \Cref{tab:hallucination_detection}. MaxProb is a competitive baseline, while Lookback Lens---which relies on attention maps derived from the generation process---is less compatible with our single-token classification setting and underperforms accordingly.

\begin{table}[t]
\centering
\resizebox{0.48\textwidth}{!}{
\begin{tabular}{l|l|ccccc}
\hline
Model & Method & WNMCQ1 & Hellaswag & GSM8K & AG News & Emotion \\
\hline
\multirow{2}{*}{LLaMA2-7B}
& MaxProb & 0.875 & \textbf{0.624} & 0.605 & 0.837 & 0.666 \\
& Lookback Lens & 0.572 & 0.530 & 0.190 & 0.572 & 0.562 \\
\hline
\end{tabular}}
\caption{AUROC ($\uparrow$) for MaxProb and Lookback Lens on hallucination detection (LLaMA2-7B), complementing \Cref{tab:hallucination_detection}. MaxProb is competitive; Lookback Lens underperforms due to incompatibility with single-token generation.}
\label{tab:hallucination_detection_additional}
\end{table}
\section{PRR Results for Hallucination Detection}\label{app:prr}

\Cref{tab:hallucination_prr} reports Prediction-Rejection Ratio (PRR) \citep{fadeeva-etal-2024-fact} for the hallucination detection benchmarks in \Cref{sec:experiments}, complementing the AUROC results in \Cref{tab:hallucination_detection}.
PRR measures the quality of an uncertainty method on the selective generation task: a higher PRR indicates that rejecting predictions with high uncertainty more reliably removes incorrect answers.
Mechanistic approaches are generally competitive with or outperform entropy-based baselines across datasets, further supporting the practical utility of self-function vectors for trustworthy generation.

\begin{table}[h]
\centering
\resizebox{0.48\textwidth}{!}{
\begin{tabular}{l|l|ccccc}
\hline
Model & Method & WNMCQ1 & Hellaswag & GSM8K & AG News & Emotion \\
\hline
\multirow{5}{*}{LLaMA2-7B}
& Total Entropy    & 0.828 & 0.242 & 0.107 & 0.776 & 0.364 \\
& Semantic Entropy & 0.815 & 0.191 & 0.277 & 0.548 & 0.283 \\
& UQ\_ICL          & 0.585 & 0.096 & 0.159 & 0.556 & 0.356 \\
& \cellcolor{gray!20}Function Vector & \cellcolor{gray!20}0.865 & \cellcolor{gray!20}\textbf{0.291} & \cellcolor{gray!20}0.518 & \cellcolor{gray!20}\textbf{0.785} & \cellcolor{gray!20}\textbf{0.374} \\
& \cellcolor{gray!20}Self-FV         & \cellcolor{gray!20}\textbf{0.866} & \cellcolor{gray!20}0.273 & \cellcolor{gray!20}\textbf{0.571} & \cellcolor{gray!20}0.779 & \cellcolor{gray!20}0.368 \\
\hline
\multirow{5}{*}{LLaMA2-13B}
& Total Entropy    & 0.841 & 0.430 & 0.179 & 0.758 & 0.386 \\
& Semantic Entropy & 0.859 & 0.394 & 0.293 & 0.536 & 0.364 \\
& UQ\_ICL          & 0.730 & 0.229 & \textbf{0.337} & \textbf{0.802} & 0.036 \\
& \cellcolor{gray!20}Function Vector & \cellcolor{gray!20}\textbf{0.875} & \cellcolor{gray!20}\textbf{0.469} & \cellcolor{gray!20}0.261 & \cellcolor{gray!20}0.783 & \cellcolor{gray!20}\textbf{0.410} \\
& \cellcolor{gray!20}Self-FV         & \cellcolor{gray!20}0.862 & \cellcolor{gray!20}0.458 & \cellcolor{gray!20}0.270 & \cellcolor{gray!20}0.777 & \cellcolor{gray!20}0.394 \\
\hline
\end{tabular}}
\caption{PRR ($\uparrow$) on diverse text classification datasets following the LM-Polygraph benchmark \citep{fadeeva-etal-2024-fact}. Mechanistic approaches are generally competitive with or outperform entropy-based baselines.}
\label{tab:hallucination_prr}
\end{table}

\end{document}